\documentclass{article}

\usepackage[final,nonatbib]{neurips_2022}

\usepackage[utf8]{inputenc} 
\usepackage[T1]{fontenc}    
\usepackage{hyperref}       
\usepackage{url}            
\usepackage{booktabs}       
\usepackage{amsfonts}       
\usepackage{nicefrac}       
\usepackage{microtype}      
\usepackage{xcolor}         
\usepackage{graphicx}

\usepackage{amsmath}
\usepackage{amsthm}
\usepackage{amsbsy,amsfonts,amssymb}
\usepackage{multirow}
\usepackage{multicol}
\usepackage{subfigure}
\usepackage{makecell}

\usepackage{enumitem}

\usepackage{hyperref}

\usepackage{cleveref}

\usepackage{algorithm}
\usepackage{algorithmic}
\usepackage[page]{appendix}
\usepackage{wrapfig}

\usepackage{titlesec}

\newcommand{\myparagraph}[1]{\noindent\textbf{#1.}}

\def\0{\textbf{0}}
\def\1{\textbf{1}}

\def\e{\boldsymbol{e}}

\def\v{\boldsymbol{v}}

\def\v{\boldsymbol{v}}

\def\x{\boldsymbol{x}}
\def\y{\boldsymbol{y}}
\def\z{\boldsymbol{z}}

\def\A{\boldsymbol{A}}

\def\cA{\mathcal{A}}

\def\xm{\boldsymbol{\xi}}

\def\zm{\boldsymbol{\zeta}}
\def\am{\boldsymbol{\alpha}}

\newcommand{\ZZ}{\bb Z} 

\renewcommand{\mathbf}{\boldsymbol}

\newcommand{\mb}{\mathbf}

\newcommand{\bb}{\mathbb}

\renewcommand{\Re}{{\mathbb R}}

\newcommand{\ours}{SDNet}

\newtheorem{theorem}{Theorem}

\title{Revisiting Sparse Convolutional Model for Visual Recognition}

\author{
\centerline{
Xili Dai\textsuperscript{\rm 1}\thanks{Equal contribution}\quad
Mingyang Li\textsuperscript{\rm 2 *}\quad
\textbf{Pengyuan Zhai}\textsuperscript{\rm 3}\quad
Shengbang Tong\textsuperscript{\rm 4}\quad
\textbf{Xingjian Gao}\textsuperscript{\rm 4} }\\
\centerline{
\textbf{Shao-Lun Huang}\textsuperscript{\rm 2}\quad
\textbf{Zhihui Zhu}\textsuperscript{\rm 5}\quad
\textbf{Chong You}\textsuperscript{\rm 4}\quad
\textbf{Yi Ma}\textsuperscript{\rm 2,4}
}\\
\centerline{
\textsuperscript{\rm 1}The Hong Kong University of Science and Technology (Guangzhou)
}\\
\centerline{ 
\textsuperscript{\rm 2}Tsinghua-Berkeley Shenzhen Institute (TBSI), Tsinghua University
}\\
\centerline{
\textsuperscript{\rm 3} Harvard University \quad
\textsuperscript{\rm 4} University of California, Berkeley \quad
\textsuperscript{\rm 5} Ohio State University
}\\
}

\begin{document}

\maketitle

\begin{abstract}
    Despite strong empirical performance for image classification, deep neural networks are often regarded as ``black boxes'' and they are difficult to interpret. On the other hand, sparse convolutional models, which assume that a signal can be expressed by a linear combination of a few elements from a convolutional dictionary, are powerful tools for analyzing natural images with good theoretical interpretability and biological plausibility. However, such principled models have not demonstrated competitive performance when compared with empirically designed deep networks. This paper revisits the sparse convolutional modeling for image classification and bridges the gap between good empirical performance (of deep learning) and good interpretability (of sparse convolutional models). Our method uses differentiable optimization layers that are defined from convolutional sparse coding as drop-in replacements of standard convolutional layers in conventional deep neural networks. We show that such models have equally strong empirical performance on CIFAR-10, CIFAR-100, and ImageNet datasets when compared to conventional neural networks. 
    By leveraging stable recovery property of sparse modeling, we further show that such models can be much more robust to input corruptions as well as adversarial perturbations in testing through a simple proper trade-off  between sparse regularization and data reconstruction terms. Source code can be found at \url{https://github.com/Delay-Xili/SDNet}.
\end{abstract}

\section{Introduction}
In recent years, deep learning has been a dominant approach for image classification and has significantly advanced the performance over previous shallow models. Despite the phenomenal empirical success, it has been increasingly realized as well as criticized that deep convolutional networks (ConvNets) are ``black boxes'' for which we are yet to develop clear understanding \cite{zhang2016understanding}.  
The layer operations such as convolution, nonlinearity and normalization are geared towards minimizing an end-to-end training loss and \emph{do not have much data-specific meaning}. 
As such, the functionality of each intermediate layer in a trained ConvNets is mostly unclear and the feature maps that they produce are hard to interpret. 
The lack of interpretability also contributes to the notorious difficulty in enhancing such learning systems for practical data which are usually corrupted by various forms of perturbation. 

This paper presents a visual recognition framework by introducing layers that \emph{have explicit data modeling} to tackle shortcomings of current deep learning systems.
We work under the assumption that the layer input can be represented by \emph{a few} atoms from a dictionary shared by all data points. 
This is the classical \emph{sparse} data modeling that, as shown in a pioneering work of \cite{olshausen1996emergence}, can easily discover meaningful structures from natural image patches.  
Backed by its ability in learning interpretable representations and strong theoretical guarantees \cite{spielman2012exact,sun2016complete,zhang2019structured,qu2019geometric,zhai2020complete,zhai2019understanding} (e.g. for handling corrupted data), sparse modeling has been used broadly in many signal and image processing applications \cite{mairal2014sparse}.
However, the empirical performance of sparse methods have been surpassed by deep learning methods for classification of modern image datasets. 

Because of the complementary benefits of sparse modeling and deep learning, there exist many efforts that leverage sparse modeling to gain theoretical insights into ConvNets and/or to develop computational methods that further improve upon existing ConvNets. 
One of the pioneering works is \cite{papyan2017convolutional} which interpreted a ConvNet as approximately solving a multi-layer convolutional sparse coding model. 
Based on this interpretation, the work \cite{papyan2017convolutional} and its follow-ups \cite{sulam2018multilayer,aberdam2019multi,zhang2021towards,cazenavette2021architectural} presented alternative algorithms and models in order to further enhance the practical performance of such learning systems. 
However, there has been no empirical evidence that such systems can handle modern image datasets such as ImageNet and obtain comparable performance to deep learning.
The only exception to the best of our knowledge is the work of \cite{sun2018supervised,sun2019supervised} which exhibited a performance on par to (on ImageNet) or better than (on CIFAR-10) ResNet. 
However, the method in \cite{sun2018supervised,sun2019supervised} 1) requires a dedicated design of network architecture that may limit its applicability, 2) is computationally orders of magnitude slower to train, and 3) does not demonstrate benefits in terms of interpretability and robustness. 
In a nutshell, sparse modeling is yet to demonstrate practicality that enables its broad applications. 


\myparagraph{Paper contributions} 
In this paper, we revisit sparse modeling for image classification and demonstrate through a simple design that sparse modeling can be combined with deep learning to obtain performance on par with standard ConvNets but with better layer-wise interpretability and stability. Our method encapsulates the sparse modeling into an \emph{implicit layer} \cite{amos2017optnet,agrawal2019differentiable,gould2019deep} and uses it as a drop-in replacement for any convolutional layer in standard ConvNets. 
The layer implements the convolutional sparse coding (CSC) model of \cite{zeiler2010deconvolutional}, and is referred to as a \emph{CSC-layer}, where the input signal is approximated by a sparse linear combination of atoms from a convolutional dictionary.
Such a convolutional dictionary is treated as the parameters of the CSC-layer that are amenable to training via back-propagation.
Then, the overall network with the CSC-layers may be trained in an end-to-end fashion from labeled data by minimizing the cross-entropy loss as usual.
This paper demonstrates that such a learning framework has the following benefits:

\begin{itemize}[leftmargin=*,topsep=0em,itemsep=0.1em]
    \item \textbf{Performance on standard datasets.} We demonstrate that our network obtains better (on CIFAR-100) or on par (on CIFAR-10 and ImageNet) performance with similar training time compared with standard architectures such as ResNet \cite{he2016deep}. 
    This provides the first evidence on the strong empirical performance of sparse modeling for deep learning to the best of our knowledge. 
    Compared to previous sparse methods \cite{sun2018supervised} that obtained similar performance, our method is of orders of magnitude faster.
    \item \textbf{Robustness to input perturbations. }
    The stable recovery property of sparse convolution model equips the CSC-layers with the ability to remove perturbation in the layer input and to recover clean sparse code. 
    As a result, our networks with CSC-layers are more robust to perturbations in the input images compared with classical neural networks. 
    Unlike existing approaches for obtaining robustness that require heavy data augmentation \cite{hendrycks2020many} or additional training techniques \cite{zheng2016improving}, our method is light-weight and does not require modifying the training procedure at all. 
\end{itemize}


\section{Related Work}

\myparagraph{Implicit layers} The idea of trainable layers defined from implicit functions can be traced back at least to the work of \cite{mairal2011task}. 
Recently, there is a revival of interests in implicit layers \cite{amos2017optnet,agrawal2019differentiable,gould2019deep,el2019implicit,bai2019deep,bai2020multiscale,wang2020implicit,liu2021convolutional} as an attractive alternative to explicit layers in existing neural networks. 
However, a majority of the cited works above define an implicit layer by a fixed point iteration, typically motivated from existing explicit layers such as residual layers, therefore they do not have clear interpretation in terms of modeling of the layer input. 
Consequently, such models do not have the ability to deal with input perturbations.
The only exceptions are differentiable optimization layers  \cite{djolonga2017differentiable,amos2017optnet,agrawal2019differentiable,amos2019differentiable} that incorporate complex dependencies between hidden layers through the formulation of convex optimization. Nevertheless, most of the above works focus on differentiating through the convex optimization layers (such as disciplined parametrized programming \cite{agrawal2019differentiable}) without specializing in any particular signal models such as the sparse models considered in this paper nor demonstrating their performance when encapsulated in multi-layer neural networks.

\myparagraph{Sparse prior in deep learning} 
Aside from image classification, sparse modeling has been introduced to deep learning for many image processing tasks such as super resolution \cite{wang2015deep}, denoising \cite{scetbon2019deep} and so on \cite{sreter2018learned,lecouat2020fully,lecouat2020flexible,liu2020interpreting}. 
These works incorporate sparse modeling by using network architectures that are \emph{motivated by} (but are not the same as) an unrolled sparse coding algorithmn LISTA \cite{gregor2010learning}. 
In sharp contrast to ours, there is no guarantee that such architectures perform a sparse encoding with respect to a particular (convolution) dictionary at all. As a result, they lack the capability of handling input perturbations as in our method. 
A notable exception is the work of \cite{sun2018supervised} where each layer performs a precise sparse encoding and exhibits on par or better performance for image classification over ResNet. 
However, the practical benefit of the sparse modeling in terms of robustness is not demonstrated. 
Moreover, \cite{sun2018supervised} adopts a patch-based sparse coding model for images and has a large computational burden.

\myparagraph{Robustness} It is known that modern neural networks are extremely vulnerable to small perturbations in the input data. A plethora of techniques have been proposed to address this instability issue, including stability training \cite{zheng2016improving}, adversarial training \cite{kurakin2016adversarial,tramer2017ensemble,athalye2018obfuscated}, data augmentation \cite{geirhos2018imagenet,yin2019fourier,hendrycks2020many}, etc. Nevertheless, these techniques either need a computational and memory overhead, or require a selection of appropriate augmentation strategies to cover all possible corruptions. With standard training only, our model can be made robust to input perturbations in test data by simply adapting sparse modeling to account for noise. Closely related to our work are \cite{gopalakrishnan2018robust,romano2020adversarial,sulam2020adversarial} which use sparse modeling to improve adversarial robustness. However, they either only demonstrate performance on very simple networks \cite{romano2020adversarial,sulam2020adversarial} or sacrifice natural accuracy for robustness \cite{gopalakrishnan2018robust}. In contrast, our method is tested on realistic networks and does not affect natural accuracy.

\section{Neural Networks with Sparse Modeling}\label{sec:network}
In this section, we show how sparse modeling is incorporated into a deep network via a specific type of network layer that we refer to as the convolutional sparse coding (CSC) layer.
We describe the CSC-layer in Sec.~\ref{sec:sc-layer} and explain how we use them for deep learning in Sec.~\ref{sec:network-architecture}.
Finally, Sec.~\ref{sec:robust-inference} explains how CSC enables robust inference with corrupted test data.

\myparagraph{Notations} Given a single-channel image $\xm \in \Re^{H \times W}$ represented as a matrix, we may treat it as a 2D signal defined on the discrete domain $[1, \ldots, H] \times [1, \ldots, W]$ and extended to $\ZZ \times \ZZ$ by padding zeros.
Given a 2D kernel $\am \in \Re^{k \times k}$, we may treat it as a 2D signal defined on the discrete domain $[-k_0\cdots, k_0] \times [-k_0 , \cdots, k_0]$
with $k = 2k_0 + 1$ and extended to $\ZZ \times \ZZ$ by padding zeros. 
Then, for convenience, we use ``$*$'' and ``$\star$'' to denote the convolution and correlation operators, respectively, between two 2D signals:
\begin{equation}
\begin{split}
      (\am * \xm)[i, j] &\doteq \sum_p \sum_q \xm[i-p, j - q] \cdot \am[p, q], \\
      (\am \star \xm)[i, j] &\doteq \sum_p \sum_q \xm[i+p, j + q] \cdot \am[p, q]. \\
\end{split}
\end{equation}

\subsection{Convolutional Sparse Coding (CSC) Layer}
\label{sec:sc-layer}

Sparse modeling is introduced in the form of an \emph{implicit layer} of a neural network. 
Unlike classical fully-connected or convolutional layers in which input-output relations are defined by an explicit function, implicit layers are defined from implicit functions. 
For our case, in particular, we introduce an implicit layer that is defined from an optimization problem involving the input to the layer as well as a weight parameter, where the output of the layer is the solution to the optimization problem. 

\myparagraph{A generative model via sparse convolution}
Concretely, given a multi-dimensional input signal $\x \in \Re^{M \times H \times W}$ to the layer where $H, W$ are spatial dimensions and $M$ is the number of channels for $\x$. We assume the signal $\x$ is generated by a multi-channel sparse code $\z \in \Re^{C \times H \times W}$ convoluting with a multi-dimensional kernel $\A \in \Re^{M \times C \times k \times k}$, which is referred to as a convolution \emph{dictionary}. Here $C$ is the number of channels for $\z$ and the convolution kernel $\A$. To be more precise, we denote $\z$ as $\z \doteq (\zm_1, \ldots, \zm_C)$ where each $\zm_c \in \Re^{H \times W}$ (presumably sparse), and denote the kernel $\A$ as
\begin{equation}\label{eq:kernel}
\resizebox{.55\hsize}{!}{$
\A \doteq
    \begin{pmatrix}
    \am_{11} & \am_{12} & \am_{13} & \dots & \am_{1C} \\
    \am_{21} & \am_{22} & \am_{23} & \dots & \am_{2C} \\
    \vdots & \vdots & \vdots & \ddots & \vdots \\
    \am_{M1} & \am_{M2} & \am_{M3} & \dots & \am_{MC}
\end{pmatrix} \quad 
\in \Re^{M \times C \times k \times k},
$}
\end{equation}
where each $\am_{ij} \in \Re^{k \times k} $ is a kernel of size $k \times k$. Then the signal $\x$ is generated via the following  operator $\cA(\cdot)$  defined by the kernel $\A$ as: 
\begin{equation}\label{eq:conv_A}
  \x =   \cA (\z) \doteq \sum_{c=1}^C \big(\am_{1c} \star \zm_c, \ldots, \am_{Mc} \star\zm_c\big) \quad \in \Re^{M \times H \times W}.
\end{equation}

\myparagraph{A layer as convolutional sparse coding} Given a multi-dimensional input signal $\x \in \Re^{M \times H \times W}$, we define that the function of ``a layer'' is to perform an (inverse) mapping to a preferably sparse output $\z_* \in \Re^{C \times H \times W}$, where $C$ is the number of output channels. Under the above sparse generative model,  we can seek the optimal sparse solution $\z$ by solving the following Lasso type optimization problem: 
\begin{equation}\label{eq:layer-def}
    \z_* = \arg\min_{\z} \lambda\|\z\|_1 + \frac{1}{2} \|\x - \cA (\z) \|_2^2 \quad \in \Re^{C \times H \times W}.
\end{equation}


\begin{wrapfigure}{r}{0.550\textwidth} 
\vspace{-.1in}
\begin{center}
\includegraphics[width=0.45\columnwidth,trim={0 2.5cm 11cm 0},clip]{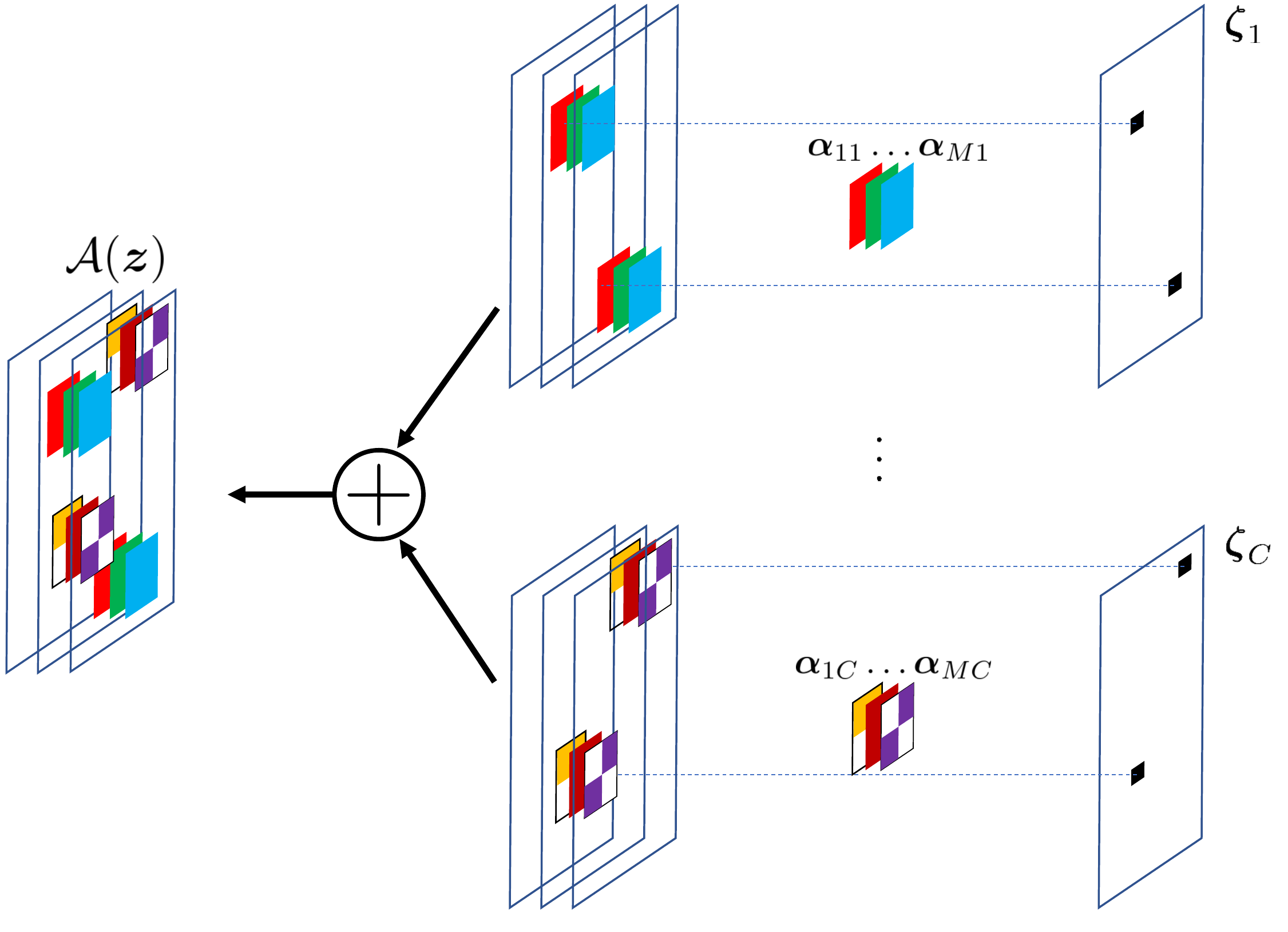}
\caption{Illustration of the operator $\cA$ in the convolutional sparse coding model for the CSC-layer. }
\label{fig:csc}
\end{center}
\vskip -0.1in
\end{wrapfigure}
The optimization problem in \eqref{eq:layer-def} is based on the convolutional sparse coding (CSC) model \cite{zeiler2010deconvolutional}\footnote{Typically, convolution operators ``$*$'' are used in the definition of the operator $\cA$ (see \eqref{eq:conv_A}), rather than the correlation operators ``$\star$''. We adopt the definition in \eqref{eq:conv_A} to be consistent with the convention of modern deep learning packages.}. 
Hence, we refer to the implicit layer defined by \eqref{eq:layer-def} as a \emph{CSC-layer}. 
The goal of the CSC model is to reconstruct the input $\x$ via $\cA (\z)$, where the feature map $\z$ specifies the locations and magnitudes of the convolutional filters in $\A$ to be linearly combined (see Figure~\ref{fig:csc} for an illustration). The reconstruction is not required to be exact in order to tolerate modeling discrepancies, and the difference between $\x$ and $\cA (\z)$ is penalized by its entry-wise $\ell_2$-norm (i.e., the $\ell_2$ norm of $\x - \cA (\z)$ flattened into a vector). 
Sparse modeling is introduced by the entry-wise $\ell_1$-norm of $\z$ in the objective function, which enforces $\z$ to be sparse. 
The parameter $\lambda>0$ controls the tradeoff between the sparsity of $\z$ and the magnitude of the residual $\x - \cA (\z)$, and is treated as a hyper-parameter that subjects to tuning via cross-validation. 
As we will show in Sec.~\ref{sec:robust-inference}, $\lambda$ can be used to improve the performance of our model in the test phase when the input is corrupted.

Based on the input-output mapping of the CSC-layer given in \eqref{eq:layer-def}, one may perform forward propagation by solving the associated optimization, and perform backward propagation by deriving the gradient of $\z_*$ with respect {{to the input $\x$}} and parameter $\A$. 
In this paper, we adopt the fast iterative shrinkage thresholding algorithm (FISTA) \cite{beck2009fast} for the forward propagation, which also produces an unrolled network architecture that can carry out automatic differentiation for backward propagation.
We defer a discussion of the implementation details of the CSC-layer to the Appendix. 


\subsection{Sparse Dictionary Learning Network Architecture and Training}
\label{sec:network-architecture}

Convolution layers are basic ingredients of ConvNets that appear in many common network architectures such as LeNet \cite{lecun1998gradient} and ResNet \cite{he2016resnet}. 
In this paper, we incorporate sparse modeling into a given existing/baseline network architecture by replacing certain / all convolution layers with the CSC-layer. Meanwhile, all other layers such as normalization, nonlinear, and fully connected layers are retained. This simple design choice allows the CSC-layers to be broadly applicable. 
We will refer to our network with CSC-layers as {\em Sparse Dictionary learning Network} (SDNet).

Give a set of training data $\{\x_i, \y_i\}_{i=1}^N$ where $\x_i$ denotes an image and $\y_i$ is the corresponding label, we train a network with CSC-layers by solving the following optimization problem:
\begin{equation}\label{eq:network-optimization}
       \min_{\mb \theta} \frac{1}{N} \sum_{i=1}^N \ell_\text{CE}\Big( f(\x_i; \;\mb \theta), \y_i \Big) ~~~~\textrm{s.t.}~~\A_s \in \mathcal{N}~~ \forall s \in \mathcal{S},
\end{equation}
where $\ell_\text{CE}$ denotes the cross-entropy loss. 
In above, we use $f(\cdot; \;\mb \theta)$ to denote the mapping that is performed by the neural network, where $\mb \theta$ is a set of learnable parameters containing a subset of kernel parameters $\{\A_s\}_{s\in \mathcal{S}}$ associated with the set $\mathcal{S}$ of CSC-layers. 
Following the convention in the sparse dictionary learning literature, we add the constraint $\A_s \in \mathcal{N}$ for the dictionaries in CSC-layers, where $\mathcal{N}$ denotes the set of normalized dictionaries:
\begin{equation*}
    \mathcal{N} \doteq \left\{\A \in \Re^{M \times C \times k \times k}:  \sum_{m=1}^M \|\am_{mc}\|_2^2 = 1, \forall c \in [C]\right\}.
\end{equation*}
To handle such a constraint, we use the projected stochastic gradient descent (SGD) for solving the problem in \eqref{eq:network-optimization}. 
That is, after each gradient update step for the parameters $\mb \theta$ and $\{\A_s\}_{s\in \mathcal{S}}$ as in a regular SGD, an extra step is taken to project each $\A_s$ onto the constraint set $\mathcal{N}$. 

\subsection{Robust Inference}
\label{sec:robust-inference}

The fundamental difference of the CSC-layer vis-a-vis a classical explicit layer (e.g., a convolutional layer) is that the CSC-layer imposes an assumption on the input feature map. 
That is, it assumes that the input feature map (or image) can be approximated by a superposition of a few atoms of a dictionary, which is the layer parameter that is learnable and is shared across all data. 
In this section, we show that CSC-layers enables us to design a robust inference strategy to obtain robustness to corruptions in ways that cannot be achieved by classical explicit layers.

We leverage an attractive property of the CSC model is that it admits a stable recovery of the sparse signals with respect to input noise. 

In particular, Theorem \ref{theorem:1} shows that \emph{any} bounded perturbation to the input of a CSC-layer produces a bounded perturbation to its output and does not change the support of the output, if one uses a $\lambda$ in \eqref{eq:layer-def} that is proportional to the norm of the perturbation in the layer input, provided that certain technical conditions are satisfied. 

\begin{theorem}(Informal version of
\cite[Theorem 19]{papyan2017working}) Suppose $\x_\natural$ has a representation $\cA (\z_\natural)$ as in \eqref{eq:conv_A}, and that it is contaminated by noise $\e$ to create the input $\x = \x_\natural + \e$. Then as long as $\z_\natural$ is sufficiently sparse, the solution $\z_*$ to \eqref{eq:layer-def} with $\lambda = O(\|\e\|_2)$ satisfies $(i)$ the support of $\z_*$ is contained in that of $\z_\natural$ and $(ii)$ $\|\z_* - \z_\natural\|_2 = O (\|\e\|_2)$.
\label{theorem:1}
\end{theorem}

\vspace{-0.5em}
Intuitively, the parameter $\lambda$ in \eqref{eq:layer-def} controls a balance between the sparsity regularization $\|\z\|_1$ and the residual $\x - \cA (\z)$, the latter of which accounts for modeling discrepancies and increases when $\x$ is noisy. 
Therefore, using a larger value of $\lambda$ helps the model to handle a larger residual. 

We thus present a very simple approach to obtain model robustness.
We consider the setting where a model is trained on an uncorrupted dataset but is tested on data that is corrupted by random noise. 
Corruptions in the input image may propagate into deeper feature maps (hence corrupting the input of all CSC-layers) during forward propagation.
Therefore, instead of directly using the CSC-layers that are obtained from the training phase, our method is to adjust the trade-off parameter $\lambda$ in the CSC-layers for the test data.
As we show in the experiments (see Sec.~\ref{sec:exp-robust}), an optimal value of $\lambda$ indeed increases with the variance of the noise in the input image, which is well-aligned with the result in Theorem \ref{theorem:1}.

\begin{algorithm}[!h]
	\caption{Robust inference with neural networks constructed from CSC-layers}
	\label{alg:robust-inference}
	\begin{algorithmic}[1]
		\REQUIRE A network architecture with CSC-layers $f(\cdot; \mb \theta, \lambda_0)$, a (clean) training data $\mathcal{T}_\text{train}$, a (corrupted) test data $\mathcal{T}_\text{test}$, corruption type $\mathtt{T}$, a set $\mathcal{C}$ of corruption levels, a set $\Lambda$ of values for $\lambda$.
		\STATE \emph{\# Training the network}
        \STATE \label{step:train} Train the network $f(\cdot; \mb \theta, \lambda_0)$ on $\mathcal{T}_\text{train}$ as described in Sec.~\ref{sec:network-architecture} to obtain learned parameters $\mb \theta_\star$.
        \STATE \emph{\# Fitting a relationship between optimal $\lambda$ and the residual from CSC-layers using $\mathcal{T}_\text{train}$}
        \FOR{each noise level $c \in \mathcal{C}$\label{step:for_noise} }
            \STATE Generate corrupted data $\mathcal{T}_\text{train}^{c}$ by injecting random noise of type $\mathtt{T}$ with level $c$ to $\mathcal{T}_\text{train}$.
            \STATE Apply $f(\cdot; \mb \theta_\star, \lambda_0)$ on $\mathcal{T}_\text{train}^{c}$ and compute averaged residual from all CSC-layers as $r_c$.
            \FOR{each parameter $\lambda \in \Lambda$}
                \STATE Apply $f(\cdot; \mb \theta_\star, \lambda)$ on $\mathcal{T}_\text{train}^{c}$ and compute averaged accuracy as $a_\lambda$.
            \ENDFOR
            \STATE Set $\lambda_c = \arg\max_{\lambda \in \Lambda} a_\lambda$.
        \ENDFOR
        \STATE \label{step:fit} Fit a function $\lambda := \lambda(r)$ from $\{\lambda_c, r_c\}_{c \in \mathcal{C}}$ via linear least squares.
        \STATE \emph{\# Computing the residual from CSC-layers on $\mathcal{T}_\text{test}$}        
        \STATE \label{step:test} Apply $f(\cdot; \mb \theta_\star, \lambda_0)$ on $\mathcal{T}_\text{test}$ and compute averaged residual from all CSC-layers as $r_\text{test}$.
		\ENSURE Predicted labels on $\mathcal{T}_\text{test}$ with the network $f(\cdot; \mb \theta_\star, \lambda(r_\text{test}))$.
	\end{algorithmic}
\end{algorithm}

\myparagraph{Choosing the optimal $\lambda$}
In practical applications the amount of noise in a given test dataset is often unknown. 
Hence, the optimal choice of parameter $\lambda$ becomes a nontrivial task. 
Here we present a practical technique for determining a proper choice of value $\lambda$, based on the simple observation that the amount of noise in the test data correlates with the magnitude of the residual $\x - \cA (\z)$.
That is, for test data corrupted by a larger amount of noise, we expect the magnitude of the residual in CSC-layers to become larger with such data fed into the network. 
Since the residual for any test data can always be computed, the key question here is how we can find a relationship between an optimal value $\lambda$ and the magnitude of the residual. 

Our technique for addressing this challenge is to learn such a relationship on the training set, by injecting synthetic data corruptions.  
For simplicity we summarize our technique in Algorithm~\ref{alg:robust-inference}, and explain it in details below. 
We assume that although the amount of noise in a test dataset $\mathcal{T}_\text{test}$ is unknown, the type of the noise (e.g., Gaussian noise, shot noise, etc.) is known. 
Given the noise type, we first determine a set $\mathcal{C}$ which contains a set of values specifying the potential amount of noise in test data.
For example, if we consider Gaussian noise, then $\mathcal{C}$ contains a set of values specifying the variance of the noise. 
We also specify a set $\Lambda$ of potential values for $\lambda$ to be used for inference. 
After training the network as specified in Step~\ref{step:train}, we use a procedure described in between Step~\ref{step:for_noise} and Step~\ref{step:fit} to fit a function that maps a residual value $r$ to an optimal choice of $\lambda$.
The idea for fitting such a relation is to generate synthetic noise of varying magnitudes in $\mathcal{C}$ on the training set, and for each noise magnitude we sweep the parameter $\lambda\in \Lambda$ to find a $\lambda$ that produces the best accuracy on training set. 
Once such a relationship is learned, we feed in the residual value computed on the test dataset $\mathcal{T}_\text{test}$ to predict an optimal $\lambda$ to be used for inference on test data, as described in Step~\ref{step:test}.

\section{Experiments}\label{sec:experiment}
In this section, we provide experimental evidence for neural networks with CSC-layers as discussed in Sec.~\ref{sec:network}. 
Through experiments on CIFAR-10, CIFAR-100\footnote{CIFAR-10 and CIFAR-100 are available at \url{https://www.cs.toronto.edu/~kriz/cifar.html}}, and ImageNet\footnote{ImageNet is a publicly available dataset: \url{https://www.image-net.org}}, Sec.~\ref{sec:image-classification} shows that our networks have equally competitive classification performance as mainstream architectures such as ResNet. 
Furthermore, we show in Sec.~\ref{sec:exp-robust} that our network is able to handle input perturbations with the robust inference technique. 
Finally, we demonstrate in Sec.~\ref{sec:exp-adversarial} that our network is able to handle adversarial perturbations as well. More details about implementation are given in the Appendix. 

\vspace{-2pt}
\myparagraph{Datasets} 
We test the performance of our method using the CIFAR-10 and CIFAR-100 \cite{krizhevsky2009learning} datasets.
Each dataset contains 50,000 training images and 10,000 testing images where each image is of size $32 \times 32$ with RGB channels. 
We also use the ImageNet dataset \cite{deng2009imagenet} that contains 1,000 classes and a total number of around 1 million images. 

\vspace{-2pt}
\myparagraph{Network architecture} 
We use the network architectures with the first convolutional layers of ResNet-18 and ResNet-34 \cite{he2016deep}\footnote{We use the implementation at \url{https://github.com/kuangliu/pytorch-cifar}, which is under the MIT License with Copyright (c) 2017 liukuang.} replaced by CSC-layers, and refer to these networks as SDNet-18 and SDNet-34, respectively. 
We use $\lambda=0.1$ as the trade-off parameter in \eqref{eq:layer-def} for all CSC-layers unless specified otherwise. 
Forward propagation through each CSC-layer is performed via unrolling two iterations of FISTA. 


\vspace{-2pt}
\myparagraph{Network training}
For CIFAR-10 and CIFAR-100, we use a cosine learning rate decay schedule with an initial learning rate of 0.1, and train the model for 220 epochs. 
We use the SGD optimizer with 0.9 momentum and Nestrov. The weight decay is set to $5 \times 10^{-4}$, and batch size is set to 128. 
All the experiments are conducted on a single NVIDIA GTX 2080Ti GPU.
For ImageNet, we use multi-step learning rate decay schedule with an initial learning rate of 0.1 decayed by a factor of $0.1$ at the 30th, 60th, and 90th epochs, and train the model for 100 epochs. The batch size is 512, and the optimizer is SGD with 0.9 momentum and Nestrov. 
All experiments on ImageNet are conducted on 4 NVIDIA RTX 3090 GPUs.

\subsection{Performance for Image Classification}\label{sec:image-classification}

\begin{table}[!t]
    \centering
    \small
    \setlength{\tabcolsep}{7.5pt}
    \renewcommand{\arraystretch}{1.1}
    \caption{ Comparison of different network archtectures, including ResNet, Multi-scale Deep Equalibrium (MDEQ), Sparse Coding Network (SCN, SCN-first), and our \ours{}, for image classification tasks. 
    We report the number of model parameters (i.e., Model Size), accuracy on test data (i.e., Top-1 Acc), GPU memory consumption during training (i.e., Memory), and the number of images that are handled per second (n/s) during training (i.e., Speed).}
    \label{tab:image-classification}
    \begin{tabular}{c c c c c c}
    \toprule
    Dataset & Architecture & Model Size & Top-1 Acc & Memory & Speed \\
    \midrule
    \multirow{6}{*}{CIFAR-10} & ResNet-18~\cite{he2016deep}   & 11.2M  & 95.54\% & 1.0 GB & 1600 n/s \\
    & ResNet-34~\cite{he2016deep}   & 21.1M  & 95.57\% & 2.0 GB & 1000 n/s \\
    & MDEQ~\cite{bai2020multiscale} & 11.1M  & 93.80\% & 2.0 GB &   90 n/s \\
    & SCN~\cite{sun2018supervised}  & 0.7M   & 94.36\% & 10.0GB &   39 n/s \\
    & SCN-18                & 11.2M  & 95.12\% & 3.5 GB &  158 n/s \\     
    & \ours-18 (ours)               & 11.2M  & 95.20\% & 1.2 GB & 1500 n/s \\
    & \ours-34 (ours)               & 21.1M  & 95.57\% & 2.4 GB &  900 n/s \\
    \midrule
    \multirow{6}{*}{CIFAR-100} & ResNet-18~\cite{he2016deep}   & 11.2M  & 77.82\% & 1.0 GB & 1600 n/s \\
    & ResNet-34~\cite{he2016deep}   & 21.1M  & 78.39\% & 2.0 GB & 1000 n/s \\
    & MDEQ~\cite{bai2020multiscale} & 11.2M  & 74.12\% & 2.0 GB &  90  n/s  \\
    & SCN~\cite{sun2018supervised}  & 0.7M   & 80.07\% & 10.0GB &  39  n/s  \\
    & SCN-18                & 11.2M  & 78.59\% & 3.5 GB &  158 n/s \\
    & \ours-18 (ours)               & 11.3M   & 78.31\% & 1.2 GB & 1500  n/s \\
    & \ours-34 (ours)               & 21.2M  & 78.48\%   & 2.4 GB & 900  n/s \\
    \midrule
    \multirow{5}{*}{ImageNet} & ResNet-18~\cite{he2016deep}   & 11.7M  & 68.98\% & 24.1 GB & 2100 n/s \\
    & ResNet-34~\cite{he2016deep}   & 21.5M  & 72.83\% & 32.3 GB & 1400 n/s \\
    & SCN~\cite{sun2018supervised}  &  9.8M  & 70.42\% & 95.1 GB & 51   n/s \\ 
    & \ours-18 (ours)               & 11.7M  & 69.47\% & 37.6 GB & 1800 n/s \\
    & \ours-34 (ours)               & 21.5M  & 72.67\% & 46.4 GB & 1200 n/s \\
    \bottomrule
    \end{tabular}
\end{table}

We compare our method with standard network architectures ResNet-18 and ResNet-34 \cite{he2016deep}.
In addition, we compare with architectures with implicit layers (i.e., MDEQ \cite{bai2020multiscale}) and architectures with sparse modeling (i.e., SCN \cite{sun2018supervised}).
For ResNet-18 and ResNet-34, we train the model using the same setup as our SDNet models.
For MDEG and SCN, we train the models using the settings as stated in their respective papers. 
SCN has both a different sparse coding layer and a different network architecture compared to our SDNet. 
Hence, we also include a baseline referred to as SCN-18, which is constructed by replacing the first convolutional layer of ResNet-18 with the sparse coding layer from SCN (hence has the same architecture as SDNet-18 but a different sparse coding layer). 


The results are reported in Table~\ref{tab:image-classification}.
We see that with a similar model size, \ours-18/34 produces a Top-1 accuracy that closely matches (for CIFAR-10 and ImageNet) or surpasses (for CIFAR-100) that of ResNet-18/34 while having a comparable Speed. 
This shows the potential of our network with modeling based layers as a powerful alternative to existing data-driven models, since our model has the additional benefit for handling corruptions. 

We also compare our \ours-18 model with the MDEQ model, which has a similar model size, and see that \ours-18 is not only more accurate than MDEQ but is much ($>7$ times) faster. 
Note that MDEQ cannot handle corrupted data as in our method as well. 

The SCN network, which also uses sparse modeling, obtains an Top-1 accuracy that is highly competitive to all methods. 
However, a significant drawback of SCN is that it is very slow to train. 
This is true even with SCN-18, where only one convolutional layer is replaced by the sparse coding layer.
The reason may be that SCN uses a patch-based sparse coding model for images, in contrast to a convolutional sparse coding model as in our method, which requires solving many sparse coding problems in each forward propagation that cannot benefit from parallel computing.



\subsection{Handling Input Perturbations}\label{sec:exp-robust}


\begin{figure}[t]
    \centering
    \subfigure[Gaussian Noise]{
        \includegraphics[width=0.23\linewidth]{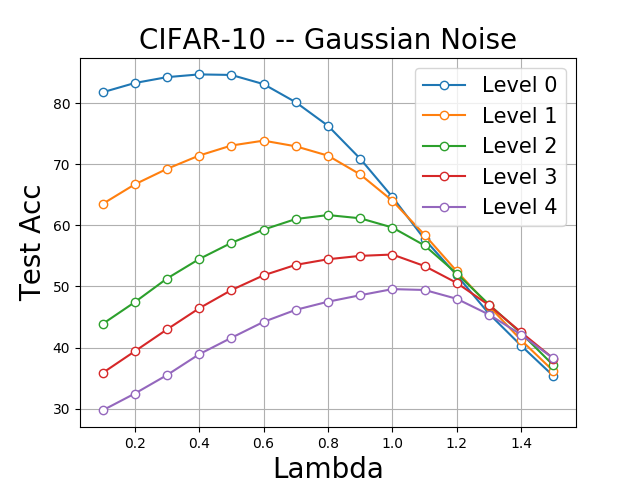}
    }
    \subfigure[Shot Noise]{
	\includegraphics[width=0.23\linewidth]{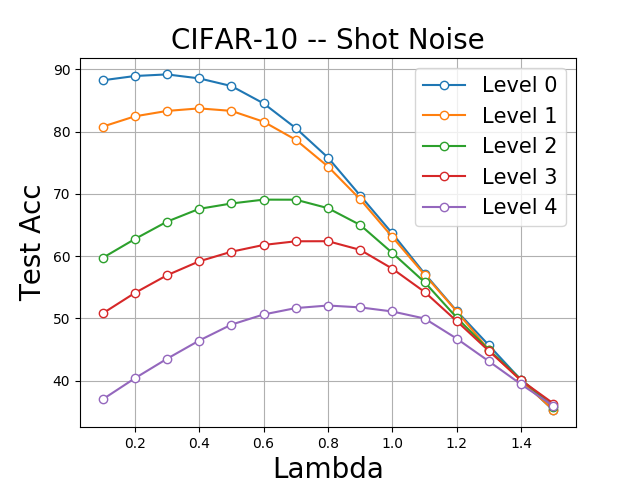}
    }
    \subfigure[Speckle Noise]{
    	\includegraphics[width=0.23\linewidth]{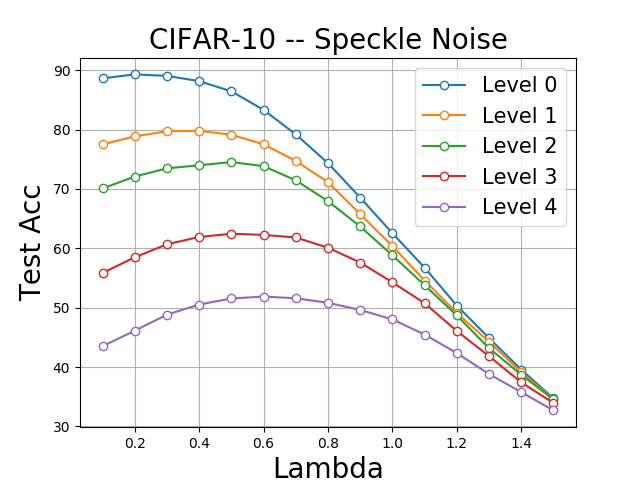}
    }
    \subfigure[Impulse Noise]{
	\includegraphics[width=0.23\linewidth]{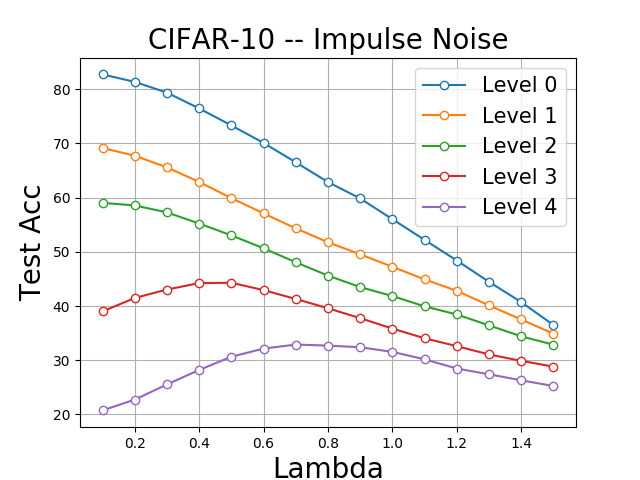}
    }
    \vspace{-3pt}
    \caption{ Test accuracy of SDNet-18 trained on CIFAR-10 dataset with $\lambda=0.1$ and evaluated on 4 types of additive noise from CIFAR-10-C \cite{hendrycks2019benchmarking} in 5 severity levels each with varying values of $\lambda$. 
    For each corruption type, optimal value of $\lambda$ for testing increases monotonically with the severity level.}
    \label{fig:robust-varying-lambda}
\end{figure}

To test the robustness of our method to input perturbations, we use the CIFAR-10-C dataset \cite{hendrycks2019benchmarking} which contains a test set for CIFAR-10 that is corrupted with different types of synthetic noise and 5 severity levels for each type. 
Because the CSC model in \eqref{eq:layer-def} penalizes the entry-wise difference between input and reconstructed signals, it is more suited for handling additive noises. 
Hence, we focus on four types of additive noises in CIFAR-10-C, namely, Gaussian noise, shot noise, speckle noise, and impulse noise.
We evaluate the accuracy of our SDNet-18 and compare its performance with ResNet-18. 

\myparagraph{Robustness as a function of $\lambda$} As discussed in Sec.~\ref{sec:robust-inference}, we may improve the performance of our model to noisy test data by using values of $\lambda$ that are different from the training phase. 
Therefore, we report the performance of our method with varying $\lambda$ in the range of $[0.1, 1.5]$ in Figure~\ref{fig:robust-varying-lambda} (recall that $\lambda=0.1$ is used for training).
It can be seen that for all types of noises and all severity levels (except for impulse noise with level 0, 1, and 2), properly choosing a value of $\lambda$ that is different from that used during training helps to improve the test performance. 
In particular, the accuracy curves as a function of $\lambda$ exhibit a unimodal shape where the performance first increases then decreases. 
Moreover, within each corruption type the values of $\lambda$ where a peak performance is achieved increases monotonically with the severity level of the corruption. 
Such an observation is well-aligned with our discussion in Sec.~\ref{sec:robust-inference}.

\myparagraph{Choosing an optimal $\lambda$} While Figure~\ref{fig:robust-varying-lambda} demonstrates that one may improve the performance on corrupted data via a proper choice of $\lambda$, it does not show how to choose the best $\lambda$ in practice. 
Here we show that the technique presented in Algorithm~\ref{alg:robust-inference} can be used to select $\lambda$ for robust inference. 
Specifically, we apply Algorithm~\ref{alg:robust-inference} with $f(\cdot; \mb \theta, \lambda_0)$ being \ours-18 with $\lambda_0 = 0.1$, $\mathcal{T}_\text{train}$ being the CIFAR-10 training set, $\mathcal{T}_\text{test}$ being CIFAR-10-C data with a particular type of corruption under a particular severity level, and report the performance of the algorithm output in Table~\ref{tab:image-classification-under-noise} and  Table~\ref{tab:image-classification-under-noise-types}.

\begin{table}[!t]
    \centering
    \small
    \setlength{\tabcolsep}{4.5pt}
    \renewcommand{\arraystretch}{1.1}
    \caption{ Classification result under varying severity levels of Gaussian noise on CIFAR-10-C with $\lambda = 0.1$ (i.e., same as training) and with an adaptive $\lambda$ computed from Algorithm~\ref{alg:robust-inference}.}
    \label{tab:image-classification-under-noise}
    \vspace{-3pt}
    \begin{tabular}{c c c c c c}
    \toprule
    Severity Level & Level-0 &  Level-1 &  Level-2 &  Level-3 &  Level-4 \\
    \midrule
    ResNet-18~\cite{he2016deep}   & 79.43\%  & 56.17\% & 34.86\% & 28.23\% & 23.45\% \\
    SCN~\cite{sun2018supervised}   & 80.89\% & 60.21\% & 44.97\% & 37.79\% & 30.11\% \\
    \ours-18 w/ $\lambda=0.1$ & 81.78\%& 63.50\% & 43.86\% & 35.84\% & 27.92\% \\
    \ours-18 w/ adaptive $\lambda$ & 84.76\%& 74.87\% & 61.38\% & 54.77\% & 48.84\% \\
    \hline

    $\lambda$ from linear fitting & 0.49 & 0.60 & 0.75 & 0.84 & 0.94 \\
    \bottomrule
    \end{tabular}
\end{table}

\begin{table}[!t]
    \centering
    \small
    \setlength{\tabcolsep}{3.5pt}
    \renewcommand{\arraystretch}{1.1}
    \caption{ Classification result under varying corruption types (averaged over all severity levels for each type) on CIFAR-10-C and ImageNet-C with  $\lambda = 0.1$ (i.e., same as training) and with an adaptive $\lambda$ computed from Algorithm~\ref{alg:robust-inference}.}
    \label{tab:image-classification-under-noise-types}
    \vspace{-3pt}
    \begin{tabular}{c c c c c c c c}
    \toprule
    & \multicolumn{4}{c}{CIFAR-10-C} & \multicolumn{3}{c}{ImageNet-C} \\ 
    Noise Type & Gaussian &  Shot &  Speckle &  Impulse & Gaussian &  Shot &  Impulse \\
    \midrule
    ResNet-18~\cite{he2016deep}   & 44.43\%  & 57.88\% & 62.16\% & 51.72\% & 22.73\% & 21.78\% & 17.38\% \\
    SCN~\cite{sun2018supervised} & 50.79\%  & 62.97\% & 67.45\% & 54.19\% & - & - & - \\
    \ours-18 w/ $\lambda=0.1$     & 50.58\%  & 63.29\% & 67.11\% & 54.13\% & 24.98\% & 23.97\% & 19.12\%  \\
    \ours-18 w/ adaptive $\lambda$& 64.92\%  & 71.13\% & 71.42\% & 57.48\% & 29.16\% & 27.59\% & 22.01\% \\
    \bottomrule
    \end{tabular}
\end{table}

Table~\ref{tab:image-classification-under-noise} shows the results with varying severity levels of Gaussian noise. 
We can see that if we use the same $\lambda = 0.1$ as in training (i.e., \ours-18 w/ $\lambda=0.1$), then the performance of \ours-18 is already significantly better ResNet-18. 
This may be attributed to the fact that the CSC-layer is intrinsically more robust to input perturbations than convolution layers owning to its sparse modeling.
However, if we use an adaptive $\lambda$ computed from Algorithm~\ref{alg:robust-inference} (i.e., \ours-18 w/ adaptive $\lambda$), then the performance is further significantly improved at all noise levels. 
To demonstrate that our method works beyond Gaussian noise, we report results for each of the four types of additive noises averaged over all severity levels in Table~\ref{tab:image-classification-under-noise-types} for CIFAR-10-C as well as ImageNet-C. 
The results demonstrate that sparse modeling enables us to effectively handle various types of additive noises in test data very easily with the procedure in Algorithm~\ref{alg:robust-inference}. 
\vspace{-3mm}
\begin{figure}[htbp]
    \centering
    \vspace{-3pt}
    \subfigure[Gaussian Noise]{
	\includegraphics[width=0.23\linewidth]{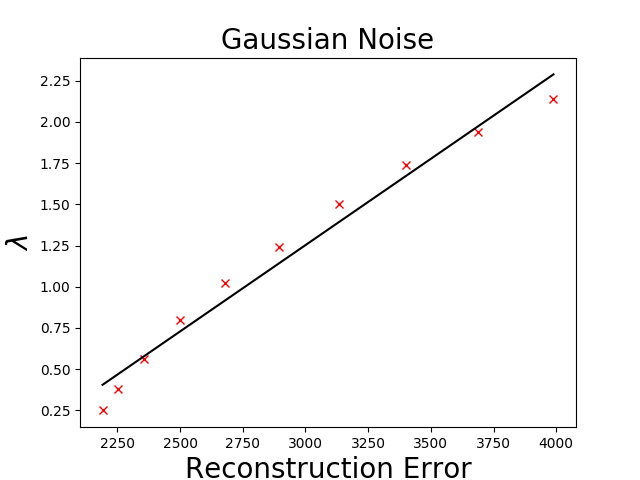}
    }
    \subfigure[Shot Noise]{
	\includegraphics[width=0.23\linewidth]{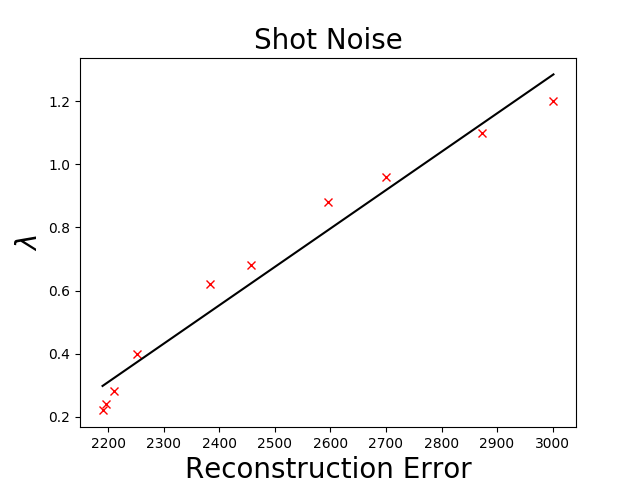}
    }
    \subfigure[Speckle Noise]{
	\includegraphics[width=0.23\linewidth]{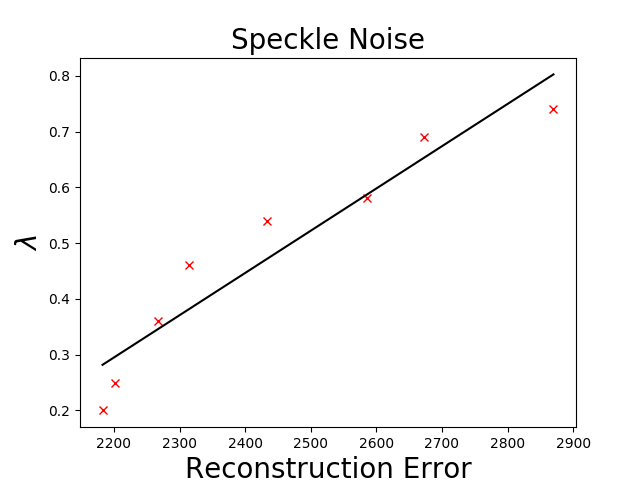}
    }
    \subfigure[Impulse Noise]{
	\includegraphics[width=0.23\linewidth]{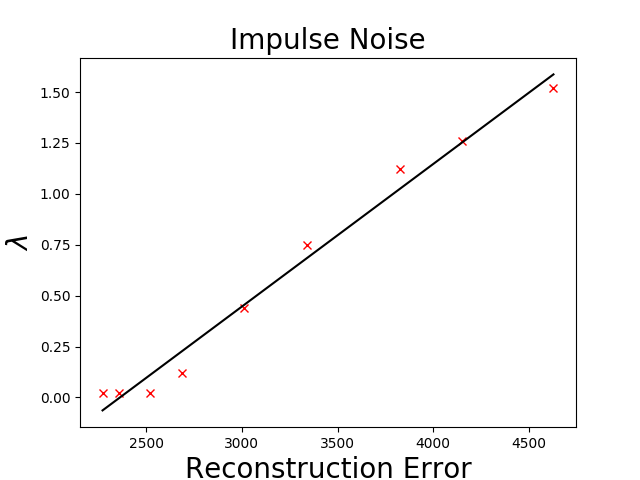}
    }
    \vspace{-5pt}
    \caption{ Relation between the optimal choice of $\lambda$ and the magnitude of residual in CSC-layers (i.e., $\{\lambda_c, r_c\}_{c \in \mathcal{C}}$ in Algorithm~\ref{alg:robust-inference}, drawn as red crosses) as well as the linear fitting (i.e., $\lambda = \lambda(c)$ in Algorithm~\ref{alg:robust-inference} , drawn as black lines) for four corruption types $\mathtt{T} \in \{$Gaussian, Shot, Speckle, Impulse$\}$.}
    \label{fig:linear-fitting}
    \vspace{-3pt}
\end{figure}

Finally, in Figure~\ref{fig:linear-fitting} we plot the relationship between the optimal choice of $\lambda$ (in y-axis) and the magnitude of residual in CSC-layers (in x-axis) learned from training data according to Algorithm~\ref{alg:robust-inference}. 
We also plot the linear fitting of such a relation, which can be seen to provide a good quality approximation for each of the four corruption types. 
On the other hand, we note that the fitted linear relations differ across different types of corruptions. 
Hence, it is important that the relationship is estimated for each corruption types separately.

\subsection{Handling Adversarial Perturbations}\label{sec:exp-adversarial}
\vspace{-2mm}
We show that our method also exhibits robustness to adversarial perturbations. 
In this experiment, we generate adversarial perturbations on the CIFAR-10 test dataset using PGD attack on our \ours~(with $\lambda=0.1$), with $L_\infty$ norm of the perturbation being $\epsilon = 8/255$ and $L_2$ norm of the perturbation being $\epsilon = 0.5$, respectively. 
The robust accuracy of our method is reported in Table~\ref{tab:image-classification-under-adversarial} and is compared with that of ResNet-18. 
We can see that while \ours~does not perform much better than ResNet with $\lambda = 0.1$, we may tune the parameter $\lambda$ to drastically improve the robust accuracy. 
\vspace{-2mm}
\begin{table}[!t]
    \centering
    \small
    \setlength{\tabcolsep}{4.5pt}
    \renewcommand{\arraystretch}{1.1}
    \caption{ Robust accuracy on CIFAR-10 with adversarial perturbation using PGD attack.}
    \label{tab:image-classification-under-adversarial}
    \begin{tabular}{c c c}
    \toprule
    Model & \makecell{Robust Accuracy \\ ($L_{\infty}=8/255$)} &  \makecell{Robust Accuracy \\ ($L_{2}=0.5$)}\\
    \midrule
    ResNet-18~\cite{he2016deep}   & 0.01\% & 29.47\% \\
    \midrule
    \ours-18 w/ $\lambda=0.1$     & 0.11\% & 29.95\% \\
    \ours-18 (After tuning $\lambda$) & 35.18\% & 62.80\%\\
    \bottomrule
    \end{tabular}
\end{table}

\subsection{Analysis}

\myparagraph{Effect of number of iterations} 
Recall from Sec.~\ref{sec:sc-layer} that forward propagation through a CSC-layer is performed by running a few iterative steps of the FISTA algorithm.
Here we provide a study on how the number of iterations affects model performance on ImageNet and ImageNet-C using SDNet-18. 
The results are shown in Table~\ref{tab:image-classification-under-noise-with-different-FISTA-iters}. 
With increasing number of FISTA iterations, the model performance on both natural accuracy and robust accuracy improves.


\begin{table}[!h]
    \centering
    \small
    \setlength{\tabcolsep}{4.5pt}
    \renewcommand{\arraystretch}{1.1}
    \caption{Effect of number of FISTA iteration on natural and robust accuracy (evaluated with ImageNet-C) for SDNet-18 trained on ImageNet.}
    \label{tab:image-classification-under-noise-with-different-FISTA-iters}
    \begin{tabular}{c c c c c}
    \toprule
    \# of FISTA Iterations  & Natural Accuracy &  Gaussian &  Shot &  Impulse \\
    \midrule
    2 & 69.47\% & 29.16\% & 27.59\% & 22.01\% \\
    4 & 69.51\% & 29.69\% & 28.15\% & 24.15\% \\
    8 & 69.79\% & 30.91\% & 29.87\% & 25.69\% \\
    \bottomrule
    \end{tabular}
\end{table}

\myparagraph{Replacing all convolution layer by the CSC-layer} 
We also train a version of \ours{} obtained from replacing all convolution layers of a ResNet with the CSC-layer (as opposed to only the first convolution layer), and refer to such a model as \ours-18-All and \ours-34-All. 
On ImageNet we observe that \ours-18-All and \ours-34-All obtain 69.37\% and 72.54\% Top-1 accuracy, respectively. 
Comparing such results with those of \ours-18/34 reported in Table~\ref{tab:image-classification}, we see that the performance is not significantly affected by replacing more convolution layers with CSC-layers (see Table~\ref{tab:the-comparison-of-SDNet-18-and-SDNet-18-All} in Appendix for more results). 
Moreover, \ours-18/34-All enables us to develop a visualization technique as described in the Appendix~\ref{sec:visualization}. 

\section{Conclusion and Discussion}

This paper revisits the classical sparse modeling and provides a simple way of using it to guide the design of interpretable deep networks. Despite multiple prior attempts, our work is the first to demonstrate that such a design can produce performance (in terms of accuracy, model size, and memory) that is on par with standard ConvNets on modern image datasets such as ImageNet. 
The success in combining sparse modeling with deep learning provides a means of borrowing and utilizing the rich results in the well-developed field of sparse modeling \cite{foucart2017mathematical,Wright-Ma-2021} for network design and analysis. While it is not the purpose of this work to fully explore all potentials, our experiments already demonstrate clear advantages of so designed networks in handling various forms of data corruptions. 
Looking forward, other fundamental principles, algorithm and techniques in sparse modeling may be introduced to further enhance the capability of our presented framework. 
Along this line, we provide some preliminary evidence on how sparse modeling enables interpretability in the Appendix, and leave further study to future work. 


\begin{ack}
Zhihui Zhu acknowledges support from NSF grants CCF-2008460. Shao-Lun Huang acknowledges support from Shenzhen Science and Technology Program under Grant KQTD20170810150821146, National Key R\&D Program of China under Grant 2021YFA0715202 and High-end Foreign Expert Talent Introduction Plan under Grant G2021032013L. Yi Ma acknowledges support from ONR grants N00014-20-1-2002 and N00014-22-1-2102, the joint Simons Foundation-NSF DMS grant \#2031899, as well as partial support from Berkeley FHL Vive Center for Enhanced Reality and Berkeley Center for Augmented Cognition, Tsinghua-Berkeley Shenzhen Institute (TBSI) Research Fund, and Berkeley AI Research (BAIR).
\end{ack}


\bibliographystyle{unsrt}

\begin{thebibliography}{10}

\bibitem{zhang2016understanding}
Chiyuan Zhang, Samy Bengio, Moritz Hardt, Benjamin Recht, and Oriol Vinyals.
\newblock Understanding deep learning requires rethinking generalization.
\newblock {\em arXiv preprint arXiv:1611.03530}, 2016.

\bibitem{olshausen1996emergence}
Bruno~A Olshausen and David~J Field.
\newblock Emergence of simple-cell receptive field properties by learning a
  sparse code for natural images.
\newblock {\em Nature}, 381(6583):607--609, 1996.

\bibitem{spielman2012exact}
Daniel~A Spielman, Huan Wang, and John Wright.
\newblock Exact recovery of sparsely-used dictionaries.
\newblock In {\em Conference on Learning Theory}, pages 37--1. JMLR Workshop
  and Conference Proceedings, 2012.

\bibitem{sun2016complete}
Ju~Sun, Qing Qu, and John Wright.
\newblock Complete dictionary recovery over the sphere i: Overview and the
  geometric picture.
\newblock {\em IEEE Transactions on Information Theory}, 63(2):853--884, 2016.

\bibitem{zhang2019structured}
Yuqian Zhang, Han-Wen Kuo, and John Wright.
\newblock {Structured Local Optima in Sparse Blind Deconvolution}.
\newblock {\em IEEE Transactions on Information Theory}, 66(1):419--452, 2019.

\bibitem{qu2019geometric}
Qing Qu, Yuexiang Zhai, Xiao Li, Yuqian Zhang, and Zhihui Zhu.
\newblock {Geometric Analysis of Nonconvex Optimization Landscapes for
  Overcomplete Learning}.
\newblock In {\em International Conference on Learning Representations}, 2019.

\bibitem{zhai2020complete}
Yuexiang Zhai, Zitong Yang, Zhenyu Liao, John Wright, and Yi~Ma.
\newblock {Complete Dictionary Learning via $\ell_{4}$-Norm Maximization over
  the Orthogonal Group.}
\newblock {\em J. Mach. Learn. Res.}, 21(165):1--68, 2020.

\bibitem{zhai2019understanding}
Yuexiang Zhai, Hermish Mehta, Zhengyuan Zhou, and Yi~Ma.
\newblock {Understanding $\ell_{4}$-based Dictionary Learning: Interpretation,
  Stability, and Robustness}.
\newblock In {\em International Conference on Learning Representations}, 2019.

\bibitem{mairal2014sparse}
Julien Mairal, Francis Bach, and Jean Ponce.
\newblock Sparse modeling for image and vision processing.
\newblock {\em arXiv preprint arXiv:1411.3230}, 2014.

\bibitem{papyan2017convolutional}
Vardan Papyan, Yaniv Romano, and Michael Elad.
\newblock Convolutional neural networks analyzed via convolutional sparse
  coding.
\newblock {\em The Journal of Machine Learning Research}, 18(1):2887--2938,
  2017.

\bibitem{sulam2018multilayer}
Jeremias Sulam, Vardan Papyan, Yaniv Romano, and Michael Elad.
\newblock Multilayer convolutional sparse modeling: Pursuit and dictionary
  learning.
\newblock {\em IEEE Transactions on Signal Processing}, 66(15):4090--4104,
  2018.

\bibitem{aberdam2019multi}
Aviad Aberdam, Jeremias Sulam, and Michael Elad.
\newblock Multi-layer sparse coding: The holistic way.
\newblock {\em SIAM Journal on Mathematics of Data Science}, 1(1):46--77, 2019.

\bibitem{zhang2021towards}
Zhiyang Zhang and Shihua Zhang.
\newblock Towards understanding residual and dilated dense neural networks via
  convolutional sparse coding.
\newblock {\em National Science Review}, 8(3):nwaa159, 2021.

\bibitem{cazenavette2021architectural}
George Cazenavette, Calvin Murdock, and Simon Lucey.
\newblock Architectural adversarial robustness: The case for deep pursuit.
\newblock In {\em Proceedings of the IEEE/CVF Conference on Computer Vision and
  Pattern Recognition}, pages 7150--7158, 2021.

\bibitem{sun2018supervised}
Xiaoxia Sun, Nasser~M Nasrabadi, and Trac~D Tran.
\newblock Supervised deep sparse coding networks.
\newblock In {\em 2018 25th IEEE International Conference on Image Processing
  (ICIP)}, pages 346--350. IEEE, 2018.

\bibitem{sun2019supervised}
Xiaoxia Sun, Nasser~M Nasrabadi, and Trac~D Tran.
\newblock Supervised deep sparse coding networks for image classification.
\newblock {\em IEEE Transactions on Image Processing}, 29:405--418, 2019.

\bibitem{amos2017optnet}
Brandon Amos and J~Zico Kolter.
\newblock Optnet: Differentiable optimization as a layer in neural networks.
\newblock In {\em International Conference on Machine Learning}, pages
  136--145. PMLR, 2017.

\bibitem{agrawal2019differentiable}
Akshay Agrawal, Brandon Amos, Shane Barratt, Stephen Boyd, Steven Diamond, and
  Zico Kolter.
\newblock Differentiable convex optimization layers.
\newblock {\em arXiv preprint arXiv:1910.12430}, 2019.

\bibitem{gould2019deep}
Stephen Gould, Richard Hartley, and Dylan Campbell.
\newblock Deep declarative networks: A new hope.
\newblock {\em arXiv preprint arXiv:1909.04866}, 2019.

\bibitem{zeiler2010deconvolutional}
Matthew~D Zeiler, Dilip Krishnan, Graham~W Taylor, and Rob Fergus.
\newblock Deconvolutional networks.
\newblock In {\em 2010 IEEE Computer Society Conference on computer vision and
  pattern recognition}, pages 2528--2535. IEEE, 2010.

\bibitem{he2016deep}
Kaiming He, Xiangyu Zhang, Shaoqing Ren, and Jian Sun.
\newblock Deep residual learning for image recognition.
\newblock In {\em Proceedings of the IEEE conference on computer vision and
  pattern recognition}, pages 770--778, 2016.

\bibitem{hendrycks2020many}
Dan Hendrycks, Steven Basart, Norman Mu, Saurav Kadavath, Frank Wang, Evan
  Dorundo, Rahul Desai, Tyler Zhu, Samyak Parajuli, Mike Guo, et~al.
\newblock The many faces of robustness: A critical analysis of
  out-of-distribution generalization.
\newblock {\em arXiv preprint arXiv:2006.16241}, 2020.

\bibitem{zheng2016improving}
Stephan Zheng, Yang Song, Thomas Leung, and Ian Goodfellow.
\newblock Improving the robustness of deep neural networks via stability
  training.
\newblock In {\em Proceedings of the ieee conference on computer vision and
  pattern recognition}, pages 4480--4488, 2016.

\bibitem{mairal2011task}
Julien Mairal, Francis Bach, and Jean Ponce.
\newblock Task-driven dictionary learning.
\newblock {\em IEEE transactions on pattern analysis and machine intelligence},
  34(4):791--804, 2011.

\bibitem{el2019implicit}
Laurent El~Ghaoui, Fangda Gu, Bertrand Travacca, and Armin Askari.
\newblock Implicit deep learning.
\newblock {\em arXiv preprint arXiv:1908.06315}, 2, 2019.

\bibitem{bai2019deep}
Shaojie Bai, J~Zico Kolter, and Vladlen Koltun.
\newblock Deep equilibrium models.
\newblock {\em Advances in Neural Information Processing Systems}, 32:690--701,
  2019.

\bibitem{bai2020multiscale}
Shaojie Bai, Vladlen Koltun, and J~Zico Kolter.
\newblock Multiscale deep equilibrium models.
\newblock {\em arXiv preprint arXiv:2006.08656}, 2020.

\bibitem{wang2020implicit}
Tiancai Wang, Xiangyu Zhang, and Jian Sun.
\newblock Implicit feature pyramid network for object detection.
\newblock {\em arXiv preprint arXiv:2012.13563}, 2020.

\bibitem{liu2021convolutional}
Sheng Liu, Xiao Li, Yuexiang Zhai, Chong You, Zhihui Zhu, Carlos
  Fernandez-Granda, and Qing Qu.
\newblock Convolutional normalization: Improving deep convolutional network
  robustness and training.
\newblock {\em Advances in Neural Information Processing Systems},
  34:28919--28928, 2021.

\bibitem{djolonga2017differentiable}
Josip Djolonga and Andreas Krause.
\newblock Differentiable learning of submodular models.
\newblock {\em Advances in Neural Information Processing Systems},
  30:1013--1023, 2017.

\bibitem{amos2019differentiable}
Brandon Amos.
\newblock {\em Differentiable optimization-based modeling for machine
  learning}.
\newblock PhD thesis, PhD thesis. Carnegie Mellon University, 2019.

\bibitem{wang2015deep}
Zhaowen Wang, Ding Liu, Jianchao Yang, Wei Han, and Thomas Huang.
\newblock Deep networks for image super-resolution with sparse prior.
\newblock In {\em Proceedings of the IEEE international conference on computer
  vision}, pages 370--378, 2015.

\bibitem{scetbon2019deep}
Meyer Scetbon, Michael Elad, and Peyman Milanfar.
\newblock Deep k-svd denoising.
\newblock {\em arXiv preprint arXiv:1909.13164}, 2019.

\bibitem{sreter2018learned}
Hillel Sreter and Raja Giryes.
\newblock Learned convolutional sparse coding.
\newblock In {\em 2018 IEEE International Conference on Acoustics, Speech and
  Signal Processing (ICASSP)}, pages 2191--2195. IEEE, 2018.

\bibitem{lecouat2020fully}
Bruno Lecouat, Jean Ponce, and Julien Mairal.
\newblock Fully trainable and interpretable non-local sparse models for image
  restoration.
\newblock In {\em European Conference on Computer Vision (ECCV) 2020}.
  Springer, 2020.

\bibitem{lecouat2020flexible}
Bruno Lecouat, Jean Ponce, and Julien Mairal.
\newblock A flexible framework for designing trainable priors with adaptive
  smoothing and game encoding.
\newblock In {\em Conference on Neural Information Processing Systems
  (NeurIPS)}, 2020.

\bibitem{liu2020interpreting}
Tianlin Liu, Anadi Chaman, David Belius, and Ivan Dokmani{\'c}.
\newblock Interpreting u-nets via task-driven multiscale dictionary learning.
\newblock {\em arXiv preprint arXiv:2011.12815}, 2020.

\bibitem{gregor2010learning}
Karol Gregor and Yann LeCun.
\newblock Learning fast approximations of sparse coding.
\newblock In {\em Proceedings of the 27th international conference on
  international conference on machine learning}, pages 399--406, 2010.

\bibitem{kurakin2016adversarial}
Alexey Kurakin, Ian Goodfellow, and Samy Bengio.
\newblock Adversarial machine learning at scale.
\newblock {\em arXiv preprint arXiv:1611.01236}, 2016.

\bibitem{tramer2017ensemble}
Florian Tram{\`e}r, Alexey Kurakin, Nicolas Papernot, Ian Goodfellow, Dan
  Boneh, and Patrick McDaniel.
\newblock Ensemble adversarial training: Attacks and defenses.
\newblock {\em arXiv preprint arXiv:1705.07204}, 2017.

\bibitem{athalye2018obfuscated}
Anish Athalye, Nicholas Carlini, and David Wagner.
\newblock Obfuscated gradients give a false sense of security: Circumventing
  defenses to adversarial examples.
\newblock In {\em International Conference on Machine Learning}, pages
  274--283. PMLR, 2018.

\bibitem{geirhos2018imagenet}
Robert Geirhos, Patricia Rubisch, Claudio Michaelis, Matthias Bethge, Felix~A
  Wichmann, and Wieland Brendel.
\newblock Imagenet-trained cnns are biased towards texture; increasing shape
  bias improves accuracy and robustness.
\newblock {\em arXiv preprint arXiv:1811.12231}, 2018.

\bibitem{yin2019fourier}
Dong Yin, Raphael~Gontijo Lopes, Jonathon Shlens, Ekin~D Cubuk, and Justin
  Gilmer.
\newblock A fourier perspective on model robustness in computer vision.
\newblock {\em arXiv preprint arXiv:1906.08988}, 2019.

\bibitem{gopalakrishnan2018robust}
Soorya Gopalakrishnan, Zhinus Marzi, Upamanyu Madhow, and Ramtin Pedarsani.
\newblock Robust adversarial learning via sparsifying front ends.
\newblock {\em arXiv preprint arXiv:1810.10625}, 2018.

\bibitem{romano2020adversarial}
Yaniv Romano, Aviad Aberdam, Jeremias Sulam, and Michael Elad.
\newblock Adversarial noise attacks of deep learning architectures: Stability
  analysis via sparse-modeled signals.
\newblock {\em Journal of Mathematical Imaging and Vision}, 62(3):313--327,
  2020.

\bibitem{sulam2020adversarial}
Jeremias Sulam, Ramchandran Muthukumar, and Raman Arora.
\newblock Adversarial robustness of supervised sparse coding.
\newblock {\em Advances in Neural Information Processing Systems},
  33:2110--2121, 2020.

\bibitem{beck2009fast}
Amir Beck and Marc Teboulle.
\newblock A fast iterative shrinkage-thresholding algorithm for linear inverse
  problems.
\newblock {\em SIAM journal on imaging sciences}, 2(1):183--202, 2009.

\bibitem{lecun1998gradient}
Yann LeCun, L{\'e}on Bottou, Yoshua Bengio, and Patrick Haffner.
\newblock Gradient-based learning applied to document recognition.
\newblock {\em Proceedings of the IEEE}, 86(11):2278--2324, 1998.

\bibitem{he2016resnet}
Kaiming He, Xiangyu Zhang, Shaoqing Ren, and Jian Sun.
\newblock Deep residual learning for image recognition.
\newblock In {\em Proceedings of the IEEE conference on computer vision and
  pattern recognition}, pages 770--778, 2016.

\bibitem{papyan2017working}
Vardan Papyan, Jeremias Sulam, and Michael Elad.
\newblock Working locally thinking globally: Theoretical guarantees for
  convolutional sparse coding.
\newblock {\em IEEE Transactions on Signal Processing}, 65(21):5687--5701,
  2017.

\bibitem{krizhevsky2009learning}
Alex Krizhevsky, Geoffrey Hinton, et~al.
\newblock Learning multiple layers of features from tiny images.
\newblock 2009.

\bibitem{deng2009imagenet}
Jia Deng, Wei Dong, Richard Socher, Li-Jia Li, Kai Li, and Li~Fei-Fei.
\newblock Imagenet: A large-scale hierarchical image database.
\newblock In {\em 2009 IEEE conference on computer vision and pattern
  recognition}, pages 248--255. Ieee, 2009.

\bibitem{hendrycks2019benchmarking}
Dan Hendrycks and Thomas Dietterich.
\newblock Benchmarking neural network robustness to common corruptions and
  perturbations.
\newblock In {\em International Conference on Learning Representations}, 2019.

\bibitem{foucart2017mathematical}
Simon Foucart and Holger Rauhut.
\newblock A mathematical introduction to compressive sensing.
\newblock {\em Bull. Am. Math}, 54(2017):151--165, 2017.

\bibitem{Wright-Ma-2021}
John Wright and Yi~Ma.
\newblock {\em High-Dimensional Data Analysis with Low-Dimensional Models:
  Principles, Computation, and Applications}.
\newblock Cambridge University Press, 2021.

\bibitem{wohlberg2015efficient}
Brendt Wohlberg.
\newblock Efficient algorithms for convolutional sparse representations.
\newblock {\em IEEE Transactions on Image Processing}, 25(1):301--315, 2015.

\bibitem{qi2020deep}
Haozhi Qi, Chong You, Xiaolong Wang, Yi~Ma, and Jitendra Malik.
\newblock Deep isometric learning for visual recognition.
\newblock In {\em International Conference on Machine Learning}, pages
  7824--7835. PMLR, 2020.

\bibitem{Erhan2009activation_max}
D.~Erhan, Yoshua Bengio, Aaron~C. Courville, and Pascal Vincent.
\newblock Visualizing higher-layer features of a deep network.
\newblock 2009.

\bibitem{Simonyan2014activation_max}
K.~Simonyan, A.~Vedaldi, and Andrew Zisserman.
\newblock Deep inside convolutional networks: Visualising image classification
  models and saliency maps.
\newblock {\em CoRR}, abs/1312.6034, 2014.

\bibitem{Mahendran_visual_network_inversion}
Aravindh Mahendran and Andrea Vedaldi.
\newblock Understanding deep image representations by inverting them.
\newblock In {\em 2015 IEEE Conference on Computer Vision and Pattern
  Recognition (CVPR)}, pages 5188--5196, 2015.

\bibitem{dosovitskiy2016inverting}
Alexey Dosovitskiy and Thomas Brox.
\newblock Inverting visual representations with convolutional networks.
\newblock In {\em Proceedings of the IEEE conference on computer vision and
  pattern recognition}, pages 4829--4837, 2016.

\bibitem{zeiler2014visualizing}
Matthew~D Zeiler and Rob Fergus.
\newblock Visualizing and understanding convolutional networks.
\newblock In {\em European conference on computer vision}, pages 818--833.
  Springer, 2014.

\end{thebibliography}

\begin{appendices}

We provide the implementation details for the CSC-layers in Appendix~\ref{sec:implementation-details-csc}, followed by a discussion on how sparse modeling enables us to easily visualize the feature maps in the intermediate layers of a convolutional neural network in Appendix~\ref{sec:visualization}.

\section{Implementation Details for CSC-Layers}
\label{sec:implementation-details-csc}

\myparagraph{Forward propagation}
Forward propagation of the sparse coding layer is carried out by solving the optimization problem in \eqref{eq:layer-def}. 
In this paper, we adopt the fast iterative shrinkage thresholding algorithm (FISTA) \cite{beck2009fast}. 
Starting with an arbitrarily initialized $\z$ (we used $\z^{[0]} = \0$), FISTA operates by introducing an auxiliary variable $\y$ initialized as $\y^{[1]} = \z^{[0]}$, as well as a scalar $m$ initialized as $m_1 = 1$, and carrying out iteratively the following steps for $\ell \ge 1$:
\begin{equation}\label{eq:ISTA}
\begin{split}
    \z^{[\ell]} &= \mathcal{T}_{\lambda t} \big( \y^{[\ell]} + t \cA^*(\x - \cA (\y^{[\ell]}) \big),\\
    m_{\ell+1} &= \frac{1 + \sqrt{1 + 4m_\ell^2}}{2}, \\
    \y^{[\ell+1]} &= \z^{[\ell]} + \frac{m_k-1}{m_{k+1}}(\z^{[\ell]} - \z^{[\ell-1]} )
\end{split}
\end{equation}
In above, $\cA^*(\cdot)$ is the adjoint operator of $\cA(\cdot)$ and it is defined as
\begin{equation}\label{eq:conv_At}
    \cA^* \x := \sum_{m=1}^M \big(\am_{m1} * \xm_m, \ldots, \am_{mC} * \xm_m\big) \in \Re^{C \times H \times W}.
\end{equation}
for any $\x = (\xm_1, \ldots, \xm_M)\in \Re^{M\times H\times W}$, where each $\xm_c \in \Re^{H \times W}$.

The FISTA iteration in \eqref{eq:ISTA} automatically gives rise to a nonlinear operator $\mathcal{T}_{\lambda t}$, which denotes a shrinkage thresholding operator that is applied entry-wise to the input variable. 
In particular, shrinkage thresholding for a scalar $\eta \in \Re$ is defined as
\begin{equation}
    \mathcal{T}_{\lambda t}(\eta) \doteq \max(|\eta| - \lambda t, 0) \cdot \textrm{sgn}(\eta).
\end{equation}

Finally, the parameter $t$ is the step size for FISTA. 
The iterations \eqref{eq:ISTA} converge to a solution to \eqref{eq:layer-def} as long as the step size $t$ is smaller than the inverse of the dominant eigenvalue of the operator $\cA^\top(\cA(\cdot))$. 
In our implementation, we estimate this dominant eigenvalue by the power iteration, which iteratively carries out the following calculation to estimate the dominant eigenvector: 
\begin{equation}\label{eq:power-iteration}
    \v^{[k+1]} \leftarrow \frac{\tilde{\v}^{[k+1]}}{\|\tilde{\v}^{[k+1]}\|_2}, ~\textrm{where}~ \tilde{\v}^{[k+1]} = \cA^\top(\cA(\v^{[k]})).
\end{equation}
The iteration is terminated when the the update $\v^{[k+1]} - \v^{[k]}$ has an entry-wise $\ell_2$-norm smaller than a predefined threshold (we use $\sqrt{10^{-5}}$) or when the maximum number of iterations (we use $50$) is reached. 
Assuming that the iteration carries out for $K$ times, the estimated dominant eigenvalue is given by
\begin{equation}
    \lambda_K = \langle \v^{[K]},  \cA^* \cA \v^{[K]} \rangle.
\end{equation}
We may then set $t$ to be a value that is smaller than $1.0 / \lambda_K$ to guarantee the convergence of ISTA. 
In our experiment we set $t = 0.9 / \lambda_K$. 
Note that the dominant eigenvalue for $\cA^\top(\cA(\cdot))$ changes after each update on the layer parameters $\A$. 
Hence, the calculation of $\lambda_K$ is performed after each parameter update during training.
Once training is completed, $\lambda_K$ can be fixed during testing and therefore does not incur additional inference time.

\myparagraph{Backward propagation}
A benefit of adopting the FISTA iteration for forward propagation in the sparse coding layer is that it naturally leads to an optimization-driven network \cite{lecouat2020flexible}, a network architecture that is constructed from an unrolled optimization algorithm, for which backward propagation can be carried out by auto-differentiation. 

We note that there are many other algorithms for solving the sparse coding and convolutional sparse coding \cite{wohlberg2015efficient} problems, which may lead to more efficient implementation of the forward propagation. 
Moreover, there are also other algorithms for performing the backward propagation \cite{mairal2011task,agrawal2019differentiable} by leveraging the fact that the sparse coding problem \eqref{eq:layer-def} is convex.
We choose FISTA because it has a very simple implementation for both forward and backward propagation. 
We leave the acceleration of our approach with other solvers as future work.

\myparagraph{Implementation}
Implementation of the sparse coding layer requires carrying out the computation of the operators $\cA(\cdot)$ and $\cA^*(\cdot)$. 
Both of these two operators can be easily implemented by many modern deep learning packages, as noted in \cite{qi2020deep}.
In PyTorch, for example, $\cA (\z)$  can be implemented by the convolutional function as follows:
\begin{equation}\label{eq:torch-conv2d}
\small  \texttt{torch.nn.functional.conv2d(input, weight)},
\end{equation}
where \texttt{input} is set to $\z$ (after inserting a batch dimension) and \texttt{weight} is set to $\A$.
Similarly, $\cA^*\x$ can be easily implemented by replacing the \texttt{conv2d} function in \eqref{eq:torch-conv2d} with the \texttt{conv\_transposed2d} function.

\section{Visualization of Feature Maps with CSC-layers}
\label{sec:visualization}

Visualization of the feature maps in the intermediate layers of a deep neural network is an important aspect toward establishing the interpretability of deep learning models. 
Here we illustrate how sparse modeling with CSC-layers enables us to very easily obtain such a visualization in ways that cannot be accomplished with standard convolutional layers. 

Before we introduce our visualization method, we mention that existing convolutional neural network visualization methods primarily include those based on 1) synthesizing an input image pattern that maximally activates one \cite{Erhan2009activation_max, Simonyan2014activation_max} or multiple \cite{Mahendran_visual_network_inversion} CNN neurons and 2) mapping the intermediate features into visually perceptible signals by training reversed convolutional filters \cite{dosovitskiy2016inverting} or by using the transposed versions of the (forward-pass) convolutional filters along with approximated unpooling operations \cite{zeiler2014visualizing}. These methods have the shortcomings such as requiring extra training steps \cite{Erhan2009activation_max, Simonyan2014activation_max, Mahendran_visual_network_inversion, dosovitskiy2016inverting}, visualizing only in the neuron level and lacking layer-level perspectives \cite{Erhan2009activation_max, Simonyan2014activation_max, zeiler2014visualizing}, and relying on numerically approximated inverse operators \cite{zeiler2014visualizing, dosovitskiy2016inverting}. 

In contrast to previous methods, our method enables an analytical reconstruction of the layer input from the layer output (feature map) by simply applying the learned convolutional dictionary on the feature map without additional training, visualization level restrictions, or any approximated inverse operators. 
As we explain next, this is made possible by sparse modeling with CSC-layers.

\subsection{Method}
\label{sec:explainability}

Recall that each CSC-layer solves an optimization problem \eqref{eq:layer-def} where the layer input is reconstructed by the layer output with layer parameter being the dictionary. 
Specifically, let $\cA_l$ be the (learned) convolutional dictionary and $\z_{l-1}$, $\z_{l}$ be the input and output feature maps respectively, at layer $\ell$. 
In CSC-layers, $\z_l$ is the optimal solution to \eqref{eq:layer-def}, replicated here for convenience:
\begin{equation}\label{eq:layer-def_at_l}
    \z_l = \arg\min_{\z} \lambda\|\z\|_1 + \frac{1}{2} \|\z_{l-1} - \cA_l (\z) \|_2^2.
\end{equation}
Given a feature map $\z_l$ at the output of a layer $\ell$, one may obtain a reconstructed layer input $\tilde{\z}_{l-1}$ from the convolution operation, i.e., $\tilde{\z}_{l-1} \doteq \cA_l(\z_l)$. 
In a deep model with multiple CSC-layers, one may apply this reconstruction step recursively from the $\ell$-th layer to the first layer, to obtain an reconstructed input image $\tilde{\x}$:
\begin{equation}\label{eq:reconstruct_step}
\begin{split}
    \tilde{\z}_{\ell-1} &\doteq \cA_n(\z_{\ell}) \\
    \tilde{\z}_{i-1} &\doteq \cA_i(\tilde{\z}_{i}),\ ~~~ i=2,\dots,\ell-1 \\
    \tilde{\x} &\doteq \cA_1(\z_1).\\ 
\end{split}
\end{equation}
To express this more compactly, we have
\begin{equation}\label{eq:reconstruct_one_line}
    \tilde{\x} \approx \cA_1(\cA_2(\dots(\cA_{\ell-1}(\cA_\ell(\z_\ell))))).
\end{equation}

The output $\tilde{\x}$, which is in the input image space, can then be visualized as an image. 
Note that the procedure in \eqref{eq:reconstruct_one_line} for obtaining $\tilde{\x}$ is analytical and thus does not require numerical approximation or additional training, as opposed to \cite{zeiler2014visualizing, dosovitskiy2016inverting} and \cite{Erhan2009activation_max, Simonyan2014activation_max, Mahendran_visual_network_inversion, dosovitskiy2016inverting}.

\begin{figure}[htbp]
    \centering
    \vspace{-6pt}
    \subfigure[input]{
	\includegraphics[width=0.46\linewidth]{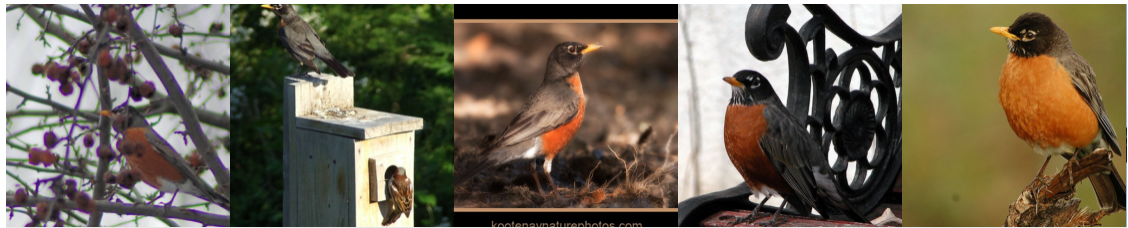}
    }
    ~
    \subfigure[layer 1]{
	\includegraphics[width=0.46\linewidth]{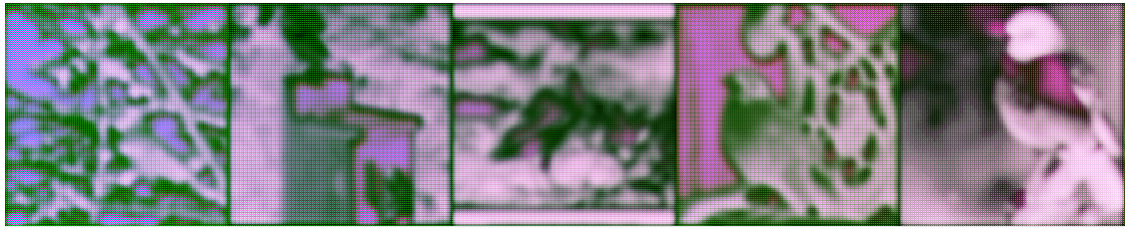}
    }
    \\
    \subfigure[layer 3]{
	\includegraphics[width=0.46\linewidth]{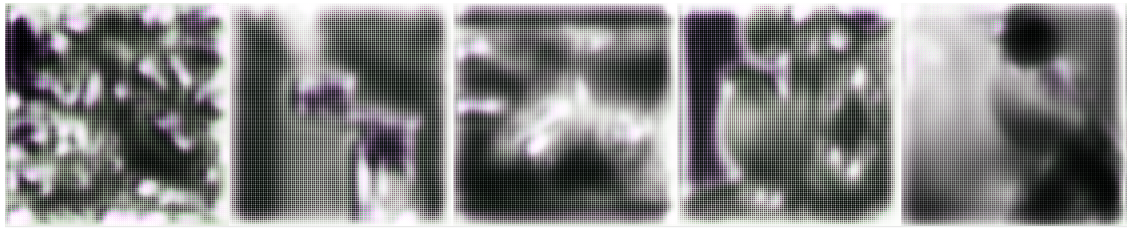}
    }
    ~
    \subfigure[layer 5]{
	\includegraphics[width=0.46\linewidth]{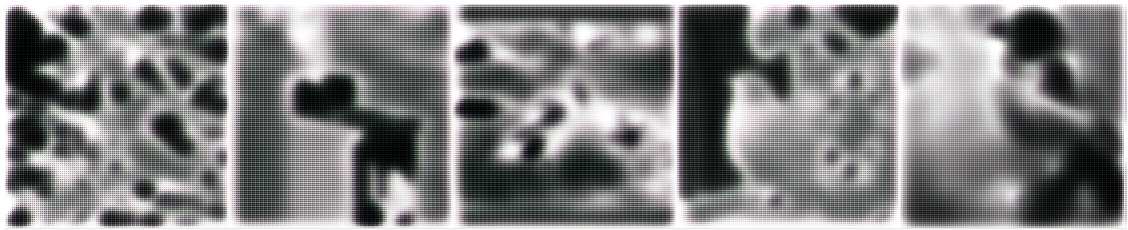}
    }
    \\
    \subfigure[layer 7]{
	\includegraphics[width=0.46\linewidth]{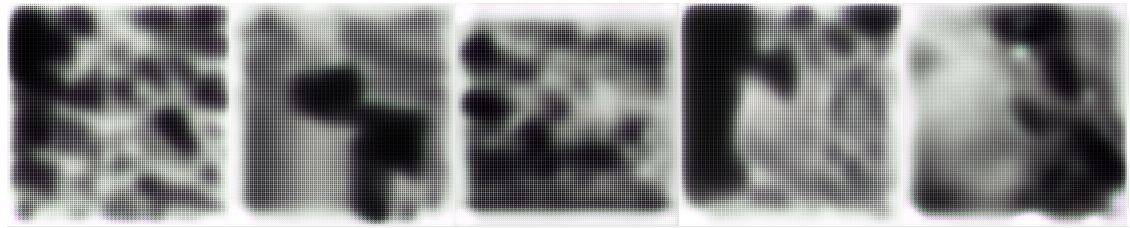}
    }
    ~
    \subfigure[layer 9]{
	\includegraphics[width=0.46\linewidth]{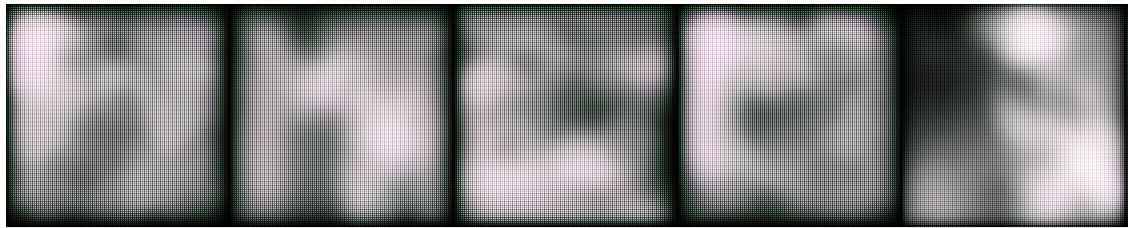}
    }
    \\
    \subfigure[layer 11]{
	\includegraphics[width=0.46\linewidth]{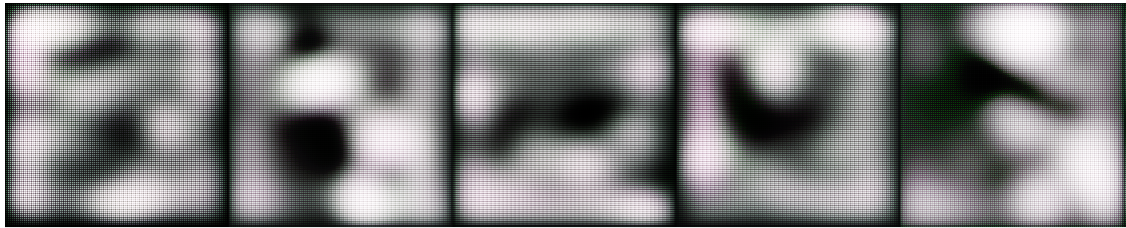}
    }
    ~
    \subfigure[layer 13]{
	\includegraphics[width=0.46\linewidth]{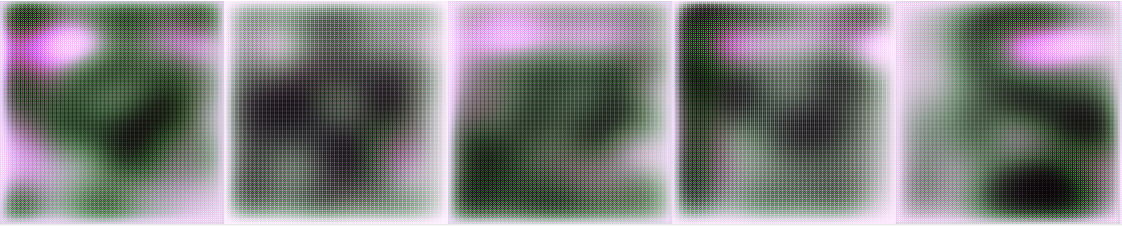}
    }
    \\
    \subfigure[layer 15]{
	\includegraphics[width=0.46\linewidth]{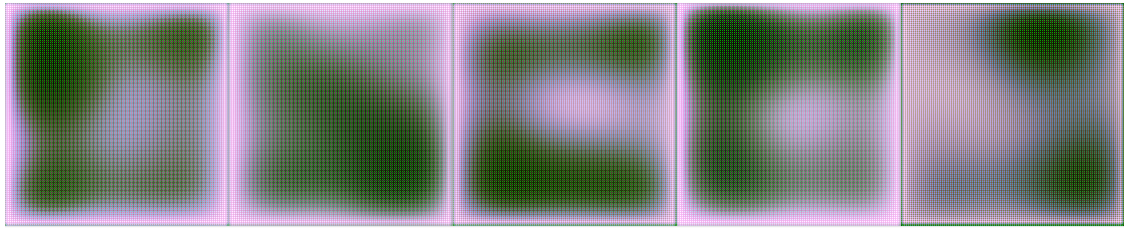}
    }
    ~
    \subfigure[layer 17]{
	\includegraphics[width=0.46\linewidth]{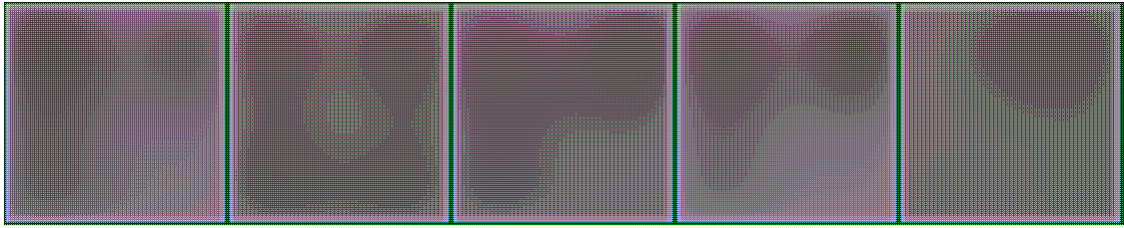}
    }
    \vspace{-5pt}
    \caption{Visualization of feature maps for 5 images at selected layers of a SDNet-18 trained on ImageNet.}
    \label{fig:revers_viz}
    \vspace{-3pt}
\end{figure}

\subsection{Results}

We apply our visualization method described above to the SDNet-18 trained on ImageNet (see Sec.~\ref{sec:experiment}) to visualize the feature maps associated with five different images. 
The results are provided in Figure~\ref{fig:revers_viz}.
It can be observed that the shallow layers (e.g., layer 1 - 5) capture rich details of the input image.
On the other hand, feature maps at deeper layers become more blurry and only capture a rough contour of the contents in the corresponding input image. 
This is suggesting that the layers of SDNet-18 progressively remove some of the unrelated details from the network input. 

\section{Visualizing the Learned Dictionary of SDNet-18}

\begin{figure}[htbp]
    \centering
    \vspace{-6pt}
	\includegraphics[width=0.46\linewidth]{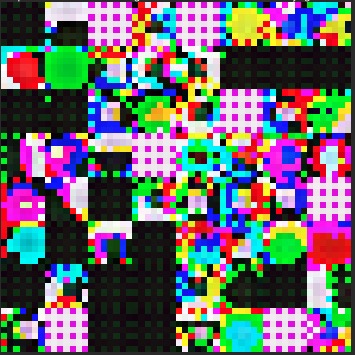}
    \vspace{-5pt}
    \caption{Visualization of the learned dictionary of first layer of SDNet-18-All trained on ImageNet.}
    \label{fig:dictionary_first_layer_sdnet18-all}
    \vspace{-3pt}
\end{figure}

In Figure~\ref{fig:dictionary_first_layer_sdnet18-all}, we provide a visualization of the learned dictionary in the first layer of SDNet-18 trained on ImageNet. The dimension of the dictionary in first layer of SDNet-18 is $64 \times 3 \times 7 \times 7$, which is treated as 64 small patches of shape $7 \times 7$ with RGB channels that are arranged into a $8 \times 8$ grid.

\section{More Experimental Results}
\label{sec:more_exp_results}



\subsection{Sparsity of CSC-layer Feature Map}
\begin{figure}[htbp]
    \centering
	\includegraphics[width=0.46\linewidth]{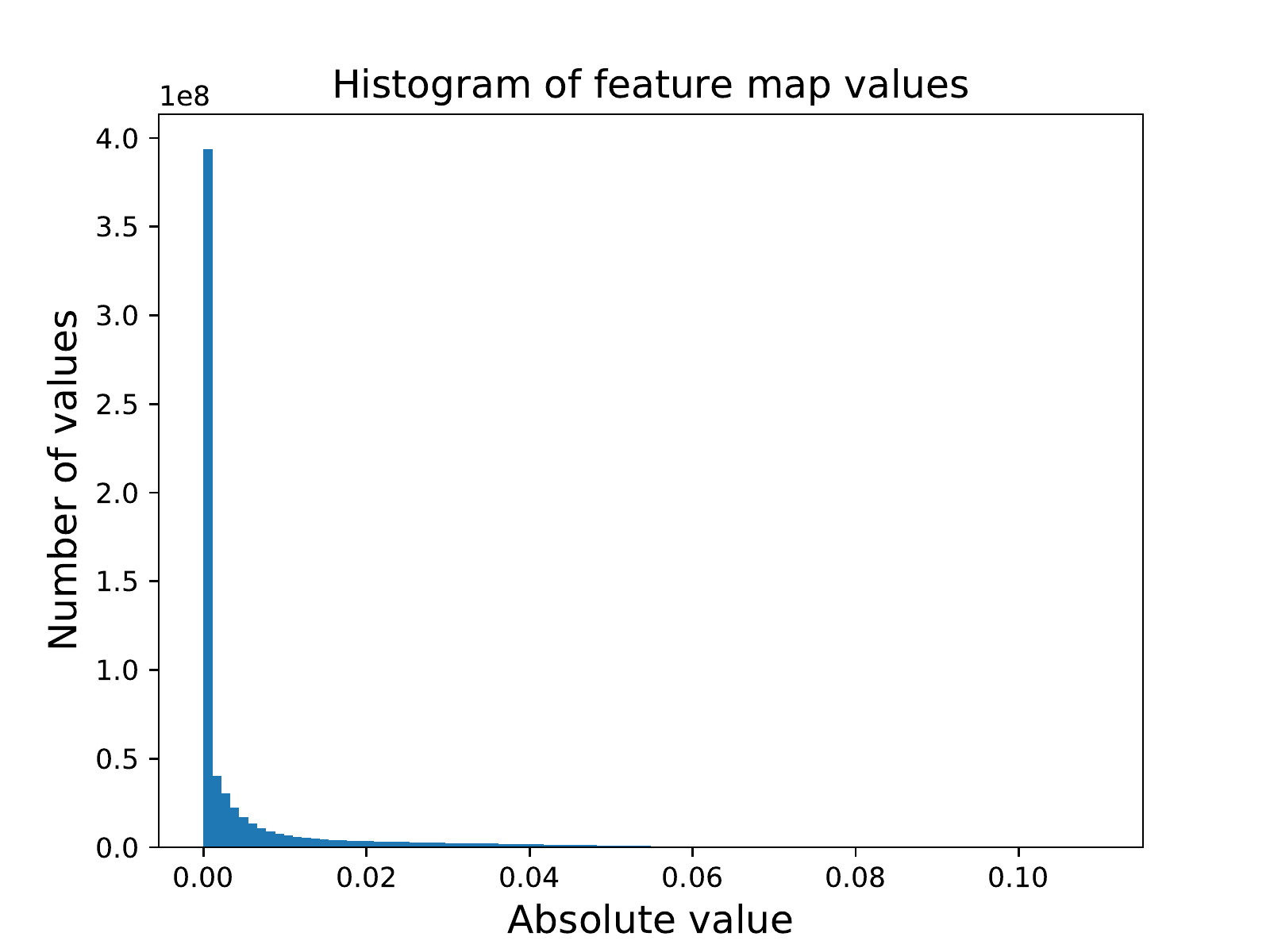}
    \caption{Histogram of CSC-layer feature map values in a trained SDNet-18 using all CIFAR-10 test images.}
    \label{fig:histogram_of_feature_map_first_layer_sdnet18-all}
\end{figure}

To evaluate the sparsity of our CSC-layer output feature map, we generate a histogram of the values at the output of CSC-layer in SDNet-18 for 10,000 CIFAR-10 test images and report the results in Figure~\ref{fig:histogram_of_feature_map_first_layer_sdnet18-all}. The result shows our CSC-layer feature maps are highly sparse. The zero value accounts for 52\% of whole CSC-layer feature maps while common convolution layer outputs dense feature maps. 

\subsection{Comparison of SDNet-18 and SDNet-18-All}
Table~\ref{tab:the-comparison-of-SDNet-18-and-SDNet-18-All} shows the comparison of SDNet-18/34 and SDNet-18/34-All on CIFAR-10, CIFAR-100 and ImageNet datasets, including accuracy, model complexity and speed. SDNet-18-All means all convolution layers are replaced with CSC-layer. And the number of FISTA iteration is set to 2 for all CSC-layers. Both models have high accuracy performance while SDNet-18 is significantly faster.

\begin{table}[!h]
    \centering
    \small
    \setlength{\tabcolsep}{4.5pt}
    \renewcommand{\arraystretch}{1.1}
    \caption{Comparison of SDNet-18 and SDNet-18-All.}
    \label{tab:the-comparison-of-SDNet-18-and-SDNet-18-All}
    \begin{tabular}{c c c c c}
    \toprule
     & Model Size &  Top-1 Acc &  Memory &  Speed \\
    \midrule
    CIFAR-10  & ~ & ~ & ~ & ~ \\
    SDNet-18     & 11.2M & 95.20\% & 1.2 GB & 1500 n/s \\
    SDNet-18-all & 11.2M & 95.18\% & 4.1 GB &  400 n/s \\
    \midrule
    CIFAR-100 & ~ & ~ & ~ & ~ \\
    SDNet-18     & 11.3M & 78.31\% & 1.2 GB & 1500 n/s \\
    SDNet-18-all & 11.3M & 78.16\% & 4.1 GB &  400 n/s \\
    \midrule
    ImageNet  & ~ & ~ & ~ & ~ \\
    SDNet-18     & 11.7M & 69.47\% & 37.6 GB & 1800 n/s \\
    SDNet-18-all & 11.7M & 69.37\% & 121.9 GB &  490 n/s \\
    SDNet-34     & 21.5M & 72.67\% & 46.4 GB & 1200 n/s \\
    SDNet-34-all & 21.5M & 72.54\% & 157.7 GB &  300 n/s \\
    \bottomrule
    \end{tabular}
\end{table}
\end{appendices}

\end{document}